\documentclass{article} 
\usepackage{iclr2025_conference,times}

\usepackage{amsmath,amsfonts,bm}

\def\eqref#1{equation~\ref{#1}}

\def\floor#1{\lfloor #1 \rfloor}
\def\1{\bm{1}}

\DeclareMathAlphabet{\mathsfit}{\encodingdefault}{\sfdefault}{m}{sl}
\SetMathAlphabet{\mathsfit}{bold}{\encodingdefault}{\sfdefault}{bx}{n}

\usepackage[hidelinks,colorlinks=true,linkcolor=blue,citecolor=blue]{hyperref}
\usepackage{pdfpages}
\usepackage{url}
\usepackage{booktabs}       
\usepackage{amsfonts}       
\usepackage{nicefrac}       
\usepackage{adjustbox}
\usepackage{microtype}      
\usepackage{xcolor} 
\usepackage{graphicx}
\usepackage{subcaption}
\usepackage{multirow}
\usepackage{amsmath}
\usepackage{enumitem}
\usepackage{listings}
\usepackage{algorithm}
\usepackage{algpseudocode}
\usepackage{tabularx}
\usepackage{adjustbox} 
\usepackage{listings}
\usepackage{tcolorbox}
\usepackage{titlecaps}
\usepackage{color,soul}

\newcommand{\method}[1]{VibeCheck}
\newcommand{\consistent}[1]{well-defined}
\newcommand{\separable}[1]{differentiating}
\newcommand{\useraligned}[1]{user-aligned}

\newcommand{\vibe}{$vibe$}

\definecolor{tan}{RGB}{254, 254, 253}

\usepackage{tcolorbox}
\tcbuselibrary{most}
\newtcolorbox{userinput}{
    enhanced,
    drop shadow,
    left=2mm,
    right=2mm,
    top=2mm,
    boxsep=1mm,
    sharp corners,
    boxrule=0pt,
    fontupper=\linespread{1.1}\small\ttfamily,
    colback=tan!2,
}
    
\newtcolorbox{modeloutput}{
    colback=gray!10,
    colframe=gray!40,
    fonttitle=\bfseries,
    coltitle=black,
    colbacktitle=gray!60,
    enhanced,
    drop shadow=black!5!white,
    left=8mm,
    right=8mm,
    top=3mm,
    bottom=3mm,
    boxsep=0mm,
    sharp corners=south,
    rounded corners=north,
    title=Model output:}

\title{\textbf{VibeCheck:} Discover \& Quantify Qualitative Differences in Large Language Models}

\iclrfinalcopy

\author{%
  Lisa Dunlap \\
  UC Berkeley\\
  \And
  Krishna Mandal \\
  UC Berkeley\\
  \And
  Trevor Darrell \\
  UC Berkeley\\
  \And
  Jacob Steinhardt \\
  UC Berkeley\\
  \And
  Joseph Gonzalez \\
  UC Berkeley\\
}

\begin{document}

\maketitle

\begin{abstract}
Large language models (LLMs) often exhibit subtle yet distinctive characteristics in their outputs that users intuitively recognize, but struggle to quantify. These "vibes" -- such as tone, formatting, or writing style -- influence user preferences, yet traditional evaluations focus primarily on the singular vibe of correctness.
We introduce \textbf{VibeCheck}, a system for automatically comparing a pair of LLMs by discovering identifying traits of a model (``vibes'') that are well-defined, differentiating, and user-aligned. VibeCheck iteratively discovers vibes from model outputs and then utilizes a panel of LLM judges to quantitatively measure the utility of each vibe. 
We validate that the vibes generated by VibeCheck align with those found in human discovery and run VibeCheck on pairwise preference data from real-world user conversations with Llama-3-70b vs GPT-4. VibeCheck reveals that Llama has a friendly, funny, and somewhat controversial vibe. These vibes predict model identity with 80\% accuracy and human preference with 61\% accuracy. Lastly, we run VibeCheck on a variety of models and tasks including summarization, math, and captioning to provide insight into differences in model behavior. \method{} discovers vibes like Command X prefers to add concrete intros and conclusions when summarizing in comparison to TNGL, Llama-405b often overexplains its thought process on math problems compared to GPT-4o, and GPT-4 prefers to focus on the mood and emotions of the scene when captioning compared to Gemini-1.5-Flash. 
Code and vibe visualizer found at \url{https://bench-mark.org/}
\end{abstract}

\section{Intro}
\label{sec:intro}

\begin{quote}
\textbf{vibe check :} A process by which a group obtains a subjective assessment of another person, place, or thing. \hfill -- \textit{Urban Dictionary}
\end{quote}

How a large language model writes a story, explains a concept, or edits an essay can be evaluated along many different dimensions such as creativity, formatting, and writing style. However, most evaluations focus on one dimension: \emph{``correctness''}. State-of-the-art in evaluation methods remain largely focused on measuring accuracy for question answering and analytical reasoning tasks~\citep{mmlu, wang2019glue, wang2019superglue, hendrycksmath2021}, and methods which aim to provide a more holistic view of LLMs~\citep{helm-instruct, padlewski2024vibeeval, mehri-eskenazi-2020-usr} rely on predefined concepts like conciseness, clarity, and trustworthiness to measure a model's performance.
These evaluation approaches fail to capture the open-ended nature of LLM applications and the critical dependence on subjective user preferences and context of the task. 
For instance, tone and creativity might be crucial in creative writing, whereas efficiency and readability are crucial in coding tasks.
To best inform users of which model would be best for their needs, we require flexible evaluation methods that can both \emph{discover} and \emph{measure} the relevant axes to evaluate for a given task.

When interacting with a set of LLMs for an extended period, a user can often tell which model generated a particular response by looking at certain traits of the outputs.
We define these identifying traits of models as \textit{``vibes''}. For instance, users have found Llama-3 outputs tend to be more friendly compared to outputs from GPT-4 and Claude which tend to be more formal (see \autoref{fig:method}); in other words, Llama-3 ranks high on the friendliness vibe, defined by the axis \texttt{formal} $\to$ \texttt{friendly}.  
Using these insights, we might select Llama for customer service tasks and Claude for coding tasks.

Understanding these vibes helps inform the development and deployment of models, but discovering and validating them for each model can be time-consuming and difficult. To address this, we outline how one can find and, more importantly, measure an LLM's vibe by formalizing three necessary and quantifiable traits of a useful vibe: \emph{\consistent{}} (agreement among multiple users), \emph{\separable{}} (ability to distinguish between models), and \emph{\useraligned{}} (predictive of user preferences).

We introduce \textbf{VibeCheck}, a system which qualitatively analyzes pairs of models by automatically finding \consistent{}, \separable{}, and \useraligned{} vibes.  
Motivated by recent work in using LLM's in lieu of human judgment \citep{llmjudge, helm-instruct, zhongd5, zhongd3, dubois2023alpacafarm}, \method{} models the qualitative analysis process by identifying the axes on which these model outputs differ to obtain a core set of vibes (e.g friendliness). 
Once these vibes are obtained, \method{} employs a panel of LLM judges~\citep{llmjury} to determine where each model's output falls on this vibe (e.g. more formal or more friendly) in order to obtain numeric scores which are then used to measure a vibe on each of our 3 key criteria. 

We run VibeCheck on several datasets to evaluate its effectiveness across different scenarios in Section~\ref{sec:results}. First, we validate that the vibes discovered by VibeCheck align well with human-annotated differences between ChatGPT and human responses using the Human ChatGPT Comparison Corpus (HC3). Next, we demonstrate that VibeCheck outperforms a predefined list of vibes in predicting user preferences on real-world comparison data from Chatbot Arena, achieving 80\% accuracy at predicting model identity and 61\% accuracy and predicting user preference. Inspecting the vibes of \method{}, we find that Llama-70b uses more typographic emphasis, more examples, and is funnier than GPT-4 and Claude-3-Opus. Conversely, we find that GPT-4 and Claude comment much more on ethics and limitations than Llama, which is more willing to give controversial responses. 

Lastly, in Section~\ref{sec:applications} we apply VibeCheck to several applications: text summarization on CNN/DailyMail, math problem-solving on MATH, and image captioning on COCO. Using \method{}, we find insightful qualitative differences between models with similar accuracy on correctness metrics but differing user preferences. For instance, Command X prefers to add concrete intros and conclusions when summarizing in comparison to TNGL, Llama-405b often overexplains its thought process on math problems, and GPT-4 prefers to focus on the mood and emotions of the scene when captioning.

\section{Related Work}
\label{sec:related}

\textbf{Aspect-based evaluations.} The number of benchmarks in the NLP community has exploded in recent years, with a growing body of work on exploring a more holistic evaluation of language models.
Several works \citep{pang2020towards, banerjee-lavie-2005-meteor, bluert} aim to improve on automatic metrics like BLEU~\citep{papineni2002bleu} and ROUGE~\citep{lin-2004-rouge} scores to better measure how well a models output aligns with the ground truth by incorporating more nuanced evaluation criteria like factual accuracy, fluency, and conciseness. Similarly, efforts have been made~\citep{helm, srivastava2023beyond, dynabench2021, wang2019glue, wang2019superglue} to standardize model evaluation by evaluating models on many of these metrics across various tasks. 

Moving away from measuring model outputs on ground truth responses, work from \citet{mehri-eskenazi-2020-usr, helm-instruct, li2019acuteeval, mehri-eskenazi-2020-unsupervised, gehrmann-etal-2021-gem} evaluate model outputs on criteria like helpfulness and clarity using LLM judges on more open ended tasks like dialogue, role-play, and summarization. While these efforts supply a great foundation for measuring correctness, they all define the axes on what makes something correct beforehand. In contrast, \method{} aims to automatically discover these axes (vibes) and verify their utility to the user by measuring the correlation between vibes and human preference.

\textbf{Pairwise comparison of LLMs.}
HCI tools like Google's AutoSxS~\citep{autosxs} and LLMComparator~\citep{llm_comparator} explores the current state of human powered LLM qualitative evaluation through interviews with data analysts. These works find that practitioners often eyeball individual examples to interpret and look at qualitative differences between the outputs of two models, and develop an interactive web based application for users to inspect side-by-side LLM outputs with an LLM based rationale as to why one output is preferred over another. While these works are focused more on software tools rather than a pipeline which can be quantitavely verified, these HCI findings inform \method{}'s vibe discovery mechanism to align with the human-powered qualitative process. Moreover, many NLP works \citep{llmjudge, llmjury, alpaca_eval, paireval2024, llmcomparative2024} have explored using LLMs to predict user preference given responses from two models, showing these preference predictions often align with the judgements of human annotators. While these efforts focus more on the user experience, it does not provide an interpretable view of exactly \textit{why} these users prefer one output over the other.

\textbf{Discovering separable traits in unstructured data.} In parallel to works in the machine learning community on LLM evaluation, there has been fantastic efforts in the HCI community on comparing generative model outputs as well as on using LLMs for qualitative analysis. Works like \citet{torii2024expanding, dispensing_humans_in_hci} use LLMs to generate discussions from qualitative research data to automate the data analysis process, but note the lack of comprehensive evaluation metrics. Automated data analysis on unstructured data has also been explored in \citet{zhongd3, zhongd5, VisDiff}, which use LLMs and VLMs to propose and validate candidate differences between two sets of text or images in the form of ``set A contains more X'', and \citet{llmmutate-24} employs an evolutionary algorithm to find text descriptions which best separates image classes to assist in zero-shot classification. We extend these works to pairwise inputs and introduce metrics of success which can better verify the separability, consistency, and alignment of these differences.

\begin{figure}[t]
    \centering
    \includegraphics[width=\linewidth, trim=0 100 0 100, clip]{figures/VibeCheck.pdf}
    \caption{\textbf{Core components of \method{}.} A vibe is an axis along which a pair of outputs differ: for example, in the top panel, output A is more friendly while output B is more formal, defining a friendliness vibe. To score a prompt output triplet, a panel of LLM judges are used to determine which output falls higher on the vibe, resulting in a score of 1 (A), -1(B), or 0(tie). Finally, the scores obtained over a large set of outputs along with preference labels are used to compute vibe utility.}
    \label{fig:method}
    \vspace{-1em}
\end{figure}
\label{sec:method}

\section{Vibe-based Evaluations}
\label{sec:formulation}

\newcommand{\pr}{\mathrm{p}}
\newcommand{\out}{\mathrm{o}}
\renewcommand{\vibe}{\nu}

We define a \emph{vibe} as an axis along which a pair of texts can differ (e.g., ``formal $\to$ friendly'') that is perceptible to humans. 
A vibe $\vibe$ is represented by a text description of the axis along with a definition of what it means to be high or low on this axis (e.g. ``Tone: low = formal, high = friendly'', see \autoref{fig:method}). Identifying vibes aids users in selecting models that best suit their specific tasks. In this work, we focus on comparing the vibes of two models by discovering the axes on which their outputs differ and quantifying the utility of these vibes.

Consider a dataset $D$ composed of triples $(\pr, \out_{A}^\pr, \out_{B}^\pr)$ and preference labels $y_\pr$, where $\pr$ is a prompt and $\out_{i}^\pr$ are the outputs from models $A$ and $B$. For each triple, a judge (human or LLM) assigns a score for vibe $\vibe$, denoted $\vibe(\pr, \out_{A}^\pr, \out_{B}^\pr) \in \{-1, 0, 1\}$, which indicates whether model $A$ scores lower (-1), similarly (0), or higher (1) than model $B$ on this vibe. Thus, a vibe imposes an ordering on model outputs.

We define 3 key criteria of a useful vibe; it should be \emph{well-defined}, \emph{\separable{}}, and \emph{user-aligned}.

\textbf{Well-defined:} multiple evaluators agree on the ordering of outputs along the vibe. We quantify this by having two different judges (typically LLMs) compute $\vibe(\pr, \out_{A}^\pr, \out_{B}^\pr)$ across dataset $D$ and report Cohen's Kappa to assess agreement.

\textbf{\titlecap{\separable{}:}} one model's outputs consistently rank higher on this vibe compared to the other's across a set of prompts. 
We quantify this by calculating a \emph{separability score} for each vibe, which measures how consistently the vibe distinguishes between the two models across all samples. 
\begin{align*}
    \mathtt{sep\_score(\vibe)} = \frac{1}{\mid D \mid}\sum_{\pr \in D}\vibe(\pr, \out_{A}^\pr, \out_{B}^\pr)
\end{align*}
To measure separability across a set of vibes, we fix a pair of models $(A, B)$ and measure the accuracy of using $\vibe(\out_{A}, \out_{B})$ to classify which output came from which model. We also more generally measure separability for a set of vibes $\nu_1, \ldots, \nu_k$, by using $\vibe_{1:k}(\pr, \out_A, \out_B)$ as a $k$-dimensional feature vector, then training a linear classifier to predict model A vs. model B, and reporting accuracy on a held-out set. We refer to this metric as \textit{model-matching accuracy}.

\textbf{User-aligned.} 
One potential use of vibes is to better understand human preferences. 
While a vibe like \emph{``frequent use of the letter `\texttt{e}' ''} may be differentiating, it is unlikely predictive of human preferences.
We assume our tuples $(\pr, \out_{A}^\pr, \out_{B}^\pr)$ are annotated with user preferences $y \in \{-1, +1\}$, indicating which model's output is preferred. We train a logistic regression classifier to predict $y$ using the same feature set $\vibe_{1:k}$ as above, reporting held-out accuracy. We refer to this metric as \emph{preference prediction accuracy}. We can measure the influence of a single vibe on preferences by examining the coefficients and p-values of the preference prediction model. 

\method{} automatically finds high-scoring vibes across the three criteria through an iterative process: (1) discovering vibes, (2) computing their scores, (3) selecting those meeting all criteria, and (4) focusing on tuples $(\pr, \out_{A}^\pr, \out_{B}^\pr)$ where existing vibes fail to differentiate the two models. 
We repeat this process to extract new, more distinguishing vibes, thus optimizing for the three key criteria while continuously refining the set of vibes.



\section{\method{}}



\method{} consists of 3 stages: vibe discovery, vibe validation, and vibe iteration. 
Further details on the method implementation and prompts used are located in the Section~\ref{sec:supp_method}.

\textbf{Vibe discovery.} Similar to how a data scientist would inspect a subset of examples to discover qualitative differences in outputs, we discover vibes by having an LLM (GPT-4o~\citep{openai2024gpt4o}) examine the differences seen in a random subset of $d$ prompt triplets. We first split the $d$ prompt triplets into smaller batches of size $batch$ and prompt GPT-4o to find differences between model $A$ and model $B$ across the set $\{(\pr_1, \out_A^1, \out_B^1), ..., (\pr_{batch}, \out_A^{batch}, \out_B^{batch})\}$. To encourage the vibes to be \consistent{} and \useraligned{}, we prompt GPT-4o to generate differences that are human-interpretable and informative for understanding the overall behaviors of A and B. Below is a paraphrased system prompt used by the proposer.


\begin{userinput}
You are a machine learning researcher analyzing outputs from two LLMs on the same input, identify differences along specific, mutually exclusive, and clearly defined axes that are easily interpretable by humans. For each axis, provide a concise description of what it means for an output to be "Low" and "High" on this axis.
\end{userinput}


 An example axis generated in this step might be `Tone: Low: formal; High: friendly'. We repeat this proposal step for $\floor{d / batch}$ sets of triplets, obtaining a final set of vibes $\{\vibe_1, .., \vibe_M\}$ by taking the union of the vibes generated in each batch. We found that GPT-4o generates 5-10 axes of variation (vibes) for each sample, so we summarize vibes across all samples in $D_{\text{discovery}}$ to find a set of $K$ vibes which appear most often in $\{\vibe_1, .., \vibe_M\}$.


\textbf{Vibe validation.} Given a vibe $\vibe$ from the discovery phase, we first apply each vibe to a set of validation tuples, then use this validation set to score vibes and compute inter-annotator agreement, model-matching accuracy, and preference prediction accuracy and filter out vibes with low scores.

To apply vibes on the validation set, we assign a score to each pair of outputs $\vibe_j(\pr, \out_{A}^\pr, \out_{B}^\pr) \in \{-1, 0, 1\}$, indicating whether model $A$ scores lower (-1), similarly (0), or higher (1) than model $B$ on the vibe. A score of 0 is assigned if the outputs are equal on this vibe or if the vibe is not applicable (e.g., the vibe is about coding style but neither output contains code); otherwise, we compute the score using a set of LLM judges (GPT-4o-mini~\citep{openai2024gpt4o} and  Llama-3-70b~\citep{llama3modelcard}). 
We average the score of the 2 judges and then round to -1, 0, or 1 (so 0.5 is rounded to 1 and -0.5 to -1). To avoid position bias~\citep{llmjudge}, we run each LLM judge twice on each sample, swapping the order of the outputs. If the judge's decision is dependent on the position of the output, we deem this pair of outputs as having a similar vibe and assign a score of $0$ for that judge.


Next, we use these scores to quantify each vibe on our 3 criteria and filter out any which are not \consistent{}, \separable{}, and \useraligned{}. 
We ensure each vibe is well-defined by computing the inter-annotator agreement (Cohen's Kappa) for each $\vibe_j$ across $D_{\text{validation}}$ and remove any with Cohen's Kappa less than 0.2, which indicates a weak agreement among judges. To ensure each vibe is \separable{}, we compute the separability score and discard any vibes with a score below $0.05$. As we explicitly prompt the model to produce vibes which provide useful insights into the behavior of language models, we assume these vibes are already aligned with users.  Using the remaining $k$ features, we run logistic regression using the scores $\vibe_{1:k}(\pr, \out_A, \out_B)$ as features to obtain our model matching and preference prediction models.



\textbf{Vibe iteration.} The filtered vibes generated in the initial vibe discovery set may not capture all the differences that contribute to user preference, resulting in a low model matching and preference prediction accuracy. We address this by 
%
iteratively refining our vibes based on tuples $(\pr, \out_{A}^\pr, \out_{B}^\pr)$ that were misclassified by our prior differentiation stages.
Specifically, we take the prompt output triplets that were misclassified by the model matching model and ask an LLM to find new axes on which these misclassified prompts vary, which are also not represented in the current set of vibes.
We then perform the same summarization/reduction procedure as before, run vibe validation/filtering, and append the resulting new vibes to the existing set of vibes.
We repeat this process for a fixed number of iterations $i$. 
In practice we find that after 3-5 iterations the discovery process does not find any additional vibes that significantly reduce the error rate of the model matching predictor. 

\section{Results}
\label{sec:results}

We first validate \method{} by comparing its discovered vibes to those identified by human annotators in Section~\ref{sec:human_vs_gpt}. Next, we evaluate \method{} on real-world user-LLM conversations with pairwise preference data, measuring how \consistent{}, \separable{}, and \useraligned{} a vibe is through inter-annotator agreement, model matching accuracy, and preference prediction accuracy on a heldout set. In Section~\ref{sec:arena} compare the discovered vibes' performance against an predefined list of common qualitative analysis criteria. Lastly, in Section~\ref{sec:applications}, we demonstrate \method{}'s broader applicability by analyzing model differences across summarization~\citep{cnn_dm}, mathematical reasoning~\citep{hendrycksmath2021}, and image captioning~\citep{coco, chen2023sharegpt4v}.



\textbf{Experimental setup.} 
Unless otherwise stated, we run \method{} for 3 iterations, use a proposer batch size of 5, and set $D_{discovery}$ to be 20 samples per iteration. Some datasets such as MATH, CNN/DailyMail, and COCO captions have no pre-computed preference labels; to simulate preferences, we apply LLM-as-a-judge and ensemble GPT-4o and Claude 3.5 Sonnet as a judge using a similar procedure to~\citep{llmjudge}, removing any samples declared a tie. Additional details on the experimental setup and hyperparameters are given in the Section~\ref{sec:exp_supp}.

We compute average Cohen's Kappa, model matching accuracy, and preference prediction accuracy on the top 10 vibes generated by \method{} on a held-out set of prompt tuples with preference labels. To obtain the top 10 vibes, we apply least-angle regression on the full set of vibes returned by \method{} to predict model identity, then sort by the separability score. The full list of vibes discovered, LR coefficients and p-values from the model matching and preference prediction models, Cohen's kappa per vibe, and separability scores are in the Section~\ref{sec:supp_vibes}. 

\textbf{List of predefined Vibes.} As a baseline, we prompt GPT-4o to generate a set of 10 vibes shown in \autoref{fig:arena_stem_writing_comparrison} and \autoref{tab:axis_definitions} which represent common axes on which LLM outputs differ. 

\subsection{Measuring \method{}'s alignment with human discovery}
\label{sec:human_vs_gpt}

We compare the findings from \method{} to findings obtained via human discovery to ensure that the vibes discovered and measured by LLM's align with humans. 
We utilize previous work~\citep{guo-etal-2023-hc3}, which collects responses written by humans and GPT-3.5~\citep{schulman2022chatgpt} for the same list of questions and then recruits 200 annotators to look at 100-200 prompt output triples presenting the characteristics of both human responses and ChatGPT answers. This results in a set of 10 insights (vibes) which are listed in detail in Section~\ref{sec:gold_standard}. 

In \autoref{tab:human_vs_gpt} we show a summarization of the top 10 vibes found by \method{} along with the corresponding insight found by humans which align with each vibe meaning. We see that \method{} uncovers most of the same vibes as the human annotators, aside from (1) GPT fabricates facts and (2) GPT focuses on a literal interpretation of the question while humans address different aspects of the question and can infer hidden meaning. The inability to find these vibes is likely a weakness of our GPT proposer, as these vibes relate to the inherent weaknesses of GPT. The complete table of \method{} outputs is located in \autoref{fig:hec3_table_overall}.

\begin{table}[h]
\vspace{-0.5em}
\small 
\setlength\tabcolsep{2pt} 
\centering
\adjustbox{max width=\textwidth}{
\begin{tabularx}{\textwidth}{X X}
\toprule
\textbf{\method{} Vibes} & \textbf{Human Discovered Vibes} \\
\midrule
\scriptsize{Humans include more references and citations} &
\scriptsize{Humans include detailed citations of papers and books.} \\
\addlinespace[1pt]
\scriptsize{GPT is more formal/academic, Humans are more casual/ conversational} &
\scriptsize{GPT answers are typically formal, humans’ are more colloquial} \\
\addlinespace[1pt]
\scriptsize{GPT includes disclaimers about advice limitations} &
\scriptsize{GPT refuses to answer questions outside its knowledge} \\
\addlinespace[1pt]
\scriptsize{GPT is cautious to give advice, emphasizes seeking professional help} &
\scriptsize{GPT shows less bias and harmful information} \\
\addlinespace[1pt]
\scriptsize{GPT has cohesive, fluid responses with clear sentence structure} &
\scriptsize{GPT writes in an organized manner with clear logic} \\
\addlinespace[1pt]
\scriptsize{GPT is strictly informative, humans include personal anecdotes} &
\scriptsize{GPT gives objective answers, humans use subjective expressions} \\
\addlinespace[1pt]
\scriptsize{GPT has less emotional engagement, humans' acknowledge emotions} &
\scriptsize{GPT expresses less emotion, humans convey their feelings} \\
\addlinespace[1pt]
\scriptsize{GPT has longer, more informative responses} &
\scriptsize{GPT has longer more detailed responses.} \\
\addlinespace[1pt]
\scriptsize{GPT has more thorough \& detailed responses} &
\scriptsize{GPT has longer more detailed responses.} \\
\addlinespace[1pt]
\scriptsize{GPT has more comprehensive responses} &
\scriptsize{GPT has longer more detailed responses.} \\

\midrule

\scriptsize{-} &
\scriptsize{GPT is strictly focused on the question, humans diverge and shift topics} \\
\addlinespace[1pt]
\scriptsize{-} &
\scriptsize{GPT may fabricate facts} \\
\bottomrule
\end{tabularx}
}
\vspace{-0.6em}
\caption{\textbf{Comparison of \method{} vibes to human labels.} Complete table in \autoref{fig:hec3_table_overall}. We see that the vibes discovered by VibeCheck closely align with vibes found through human analysis.}
\vspace{-1.5em}
\label{tab:human_vs_gpt}
\end{table}

\newcolumntype{L}[1]{>{\raggedright\arraybackslash}p{#1}}

\subsection{Describing user preference on Chatbot Arena}
\label{sec:arena}
 
On April 18th 2024, Meta released their open-weight large language model Llama 3. On Chatbot Arena~\citep{chiang2024chatbot}, a popular human preference benchmark, Llama-3-70b achieves similar preference scores to models like GPT-4 and Claude 3 Opus, despite these models outperforming Llama on traditional benchmarks like MMLU. This has led to speculation on whether there are qualitative properties of Llama that make it popular among users~\citep{llama3arena2024}.

On Chatbot Arena, we run \method{} on a set of combined battles (pairwise comparisons) between Llama-3-70b VS GPT-4 and Llama-3-70b VS Claude3-Opus\footnote{Data:  \url{https://huggingface.co/datasets/lmarena-ai/Llama-3-70b-battles}} under three settings: using the entire dataset, and using 2 subsets of the data: STEM prompts (including coding) and  Writing prompts, which include creative writing, humanities questions, and general chatting. We obtain these subsets by using GPT-4o-mini to categorize the questions as a STEM Question, a Writing/Chatting prompt, or neither. The size of each subset can be found in Section~\ref{sec:exp_supp}.

\begin{figure}[h]
    \centering
    \includegraphics[width=\linewidth, trim={0.4cm 18.3cm 0.4cm 0.5cm},clip]{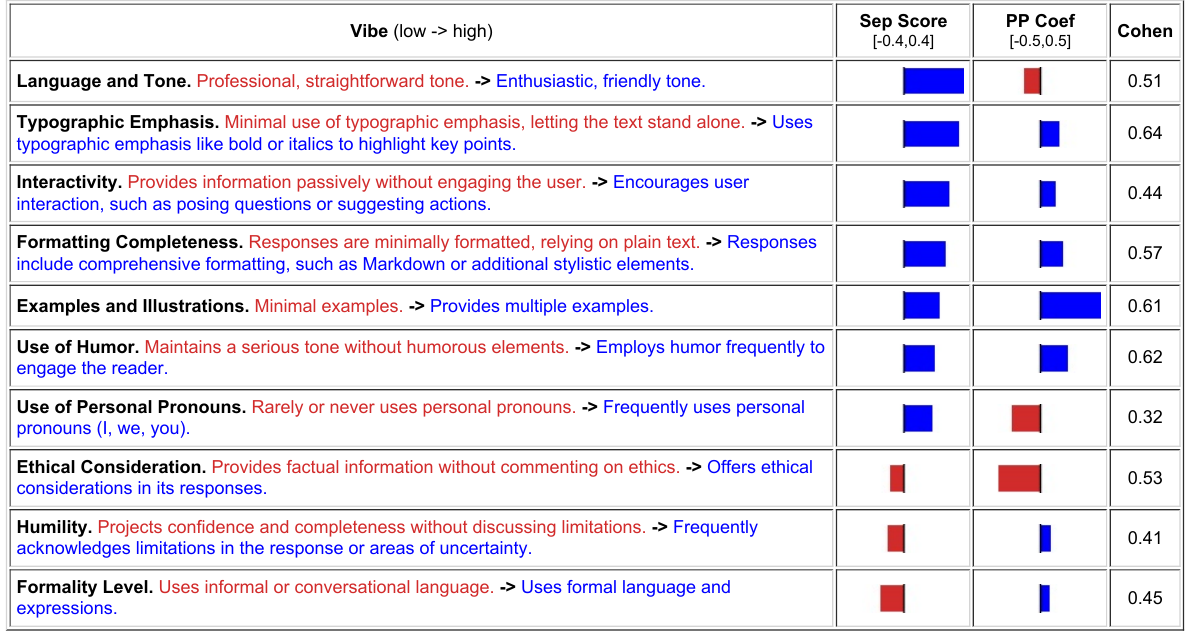}
    \vspace{-2em}
    \caption{\textbf{Comparing Llama-3-70b VS GPT-4 \& Claude-3-Opus on Chatbot Arena.} Negative separability scores indicate Llama-3-70B aligns with the low (red) description, while negative preference coefficients show alignment with low descriptions is preferred. We see that Llama is more humorous, utilizes more formatting, provides more examples, and comments much less on ethics than GPT and Claude: all attributes which correlate positively with human preference.}
    \vspace{-1.3em}
    \label{fig:arena_overall_3_summary}
\end{figure}

We compare the vibes found by \method{} to a list of predefined vibes (\autoref{tab:axis_definitions}) of common differences between language models which a user may be interested in. \autoref{tab:arena_vibe_results} shows that \method{} achieves higher model matching accuracy than the predefined vibes all categories and more iterations improve model matching and preference prediction accuracy. Furthermore, \autoref{fig:arena_overall_3_summary} shows that the vibes are more fine-grained. We summarize our other findings below:

\textbf{Comparing MM and PP accuracy across topics.} \autoref{tab:arena_vibe_results} shows that MM and PP accuracy is lower for STEM questions compared to writing or overall prompts. We suspect this is because Llama’s qualitative traits (friendliness, humor, safety, etc.) are less relevant for objective questions like coding and math, and user preferences here are influenced more by factual accuracy than stylistic traits. Conversely, \method{} best predicts preferences for writing-oriented prompts, as style is often more important for these open ended tasks.

\begin{figure}[h]
    \centering
    \includegraphics[width=\linewidth, trim={0.4cm 17.5cm 0.4cm 0.5cm},clip]{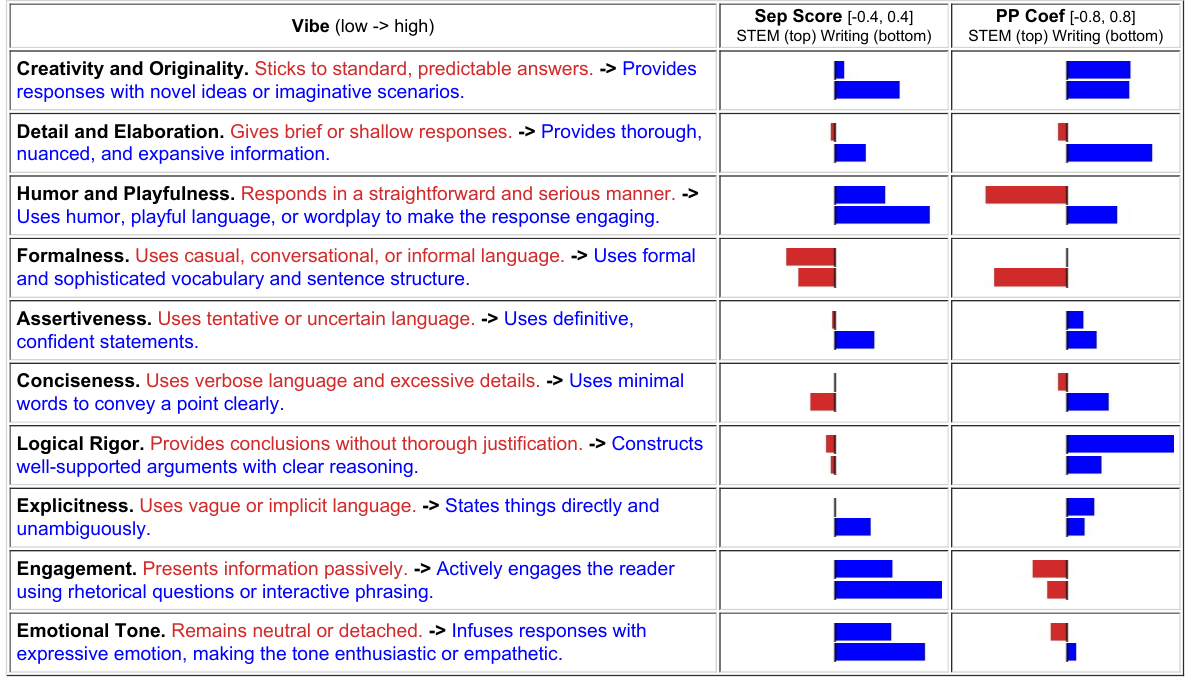}
    \vspace{-2.3em}
    \caption{\textbf{Comparing user preference and separability across STEM and writing tasks.} On predefined list of vibes referenced in \autoref{tab:arena_vibe_results}. Negative preference coefficients indicate a preference for low-description vibes, while negative separability scores show Llama responses align more with the low description than Claude or GPT responses. For writing tasks, detailed explanations, humor, and expressive emotion positively correlate with human preference, while these traits negatively correlate with STEM tasks. Conversely, logical rigor has a stronger positive impact on preference for STEM tasks. These trends are reflected in separability scores, with less separability on STEM tasks for vibes like humor and emotional tone, and more separability for logical rigor.}
    \vspace{-2em}
    \label{fig:arena_stem_writing_comparrison}
\end{figure}

To understand how user preferences for these vibes vary across task domains and contexts, we analyze separability scores and preference prediction coefficients for predefined vibes in \autoref{fig:arena_stem_writing_comparrison}. For writing tasks, formality, humor, and expressive emotional content positively correlate with user preference, while these traits negatively correlate with STEM tasks, where logical rigor is the most influential on preference. Conversely, logical rigor has minimal impact on preferences for writing tasks. While our dataset does not directly compare individual judgments, treating STEM and writing task users as distinct groups provides preliminary evidence of task-specific preferences. Additionally, lower separability scores for STEM tasks indicate less stylistic divergence in model outputs for objective questions like coding and math, making model identity harder to predict, consistent with \autoref{tab:arena_vibe_results}.

\textbf{Notable vibes for Llama-3 70B.} The top 10 vibes uncovered by \method{} (\autoref{fig:arena_overall_3_summary}) highlight Llama’s use of formatting, willingness to engage with sensitive topics, less emphasis on ethics, and a conversational, humorous style. Finer-grained vibes include Llama’s use of bold/italics to emphasize points and increased use of personal pronouns, with ‘I,’ ‘we,’ and ‘you’ appearing ~3x more in Llama outputs than GPT/Claude conversations. The preference prediction coefficeients in \autoref{fig:arena_overall_3_summary} show Chatbot Arena users tend to prefer outputs which are less focused on ethics, employ markdown and typographic emphasis to highlight key points, and employ humor to engage the user, all of which are vibes which llama possesses.  We believe that this correlation between vibes and user preference can explain some of the discrepancy seen in llamas high ranking on the leaderboard in comparison to models like GPT-4 which often outperform Llama.

\begin{table}[h]
\centering
\begin{tabular}{l|ccc|ccc|ccc}
\toprule
\multicolumn{1}{c|}{Method} & \multicolumn{3}{c|}{Overall} & \multicolumn{3}{c|}{STEM} & \multicolumn{3}{c}{Writing} \\
\cmidrule(lr){2-4} \cmidrule(lr){5-7} \cmidrule(lr){8-10}
 & M.M. & P.P. & C.K. & M.M. & P.P. & C.K. & M.M. & P.P. & C.K. \\
\midrule
\method{} [1 iter] & 68.68 & 60.00 & 0.42 & 65.20 & 55.99 & 0.44 & 74.09 & 60.58 & 0.51\\
\method{} [3 iter] & \textbf{80.34} & 59.34 & 0.46 & \textbf{68.71} & 57.31 & 0.45 & \textbf{77.19} & \textbf{62.04} & 0.49\\
\midrule
Predefined Vibes & 72.10 & \textbf{61.11} & 0.51 & 65.94 & \textbf{58.38} & 0.45 & 75.00 & 59.49 & 0.52 \\
\bottomrule
\end{tabular}
\vspace{-0.5em}
\caption{\textbf{Comparing Llama-3 to GPT and Claude on Chatbot Arena.} We report Model Matching Accuracy (M.M.), Preference Prediction Accuracy (P.P.), and average Cohen’s Kappa (C.K) for the full dataset (Overall) and STEM and Writing categories. \method{} achieves higher model matching accuracy than Predefined Vibes and similar preference prediction accuracy. \method{} obtains the largest improvements over predefined vibes in the writing category, suggesting that for open-ended prompts, model styles differ significantly, and style has a greater influence on preference.}
\vspace{-1em}
\label{tab:arena_vibe_results}
\end{table}

\section{Applications}
\label{sec:applications}

We next apply VibeCheck to discover qualitative differences between models' behavior on three open-ended tasks: text summarization, math problem-solving, and image captioning. We use CNN/DailyMail~\citep{cnn_dm} for text summarization, MATH~\citep{hendrycks2021measuring} with chain-of-thought prompting for problem-solving, and COCO for image captioning. For CNN and MATH we use cached model predictions downloaded from HELM~\citep{helm} and intentionally choose models which are ranked similarly to each other, but when running LLM as a judge to get preference labels, one model is more heavily preferred. For captioning, we generate captions on a random subset of 1000 COCO images. The vibes for each application in Section~\ref{sec:supp_vibes}. 
\subsection{What do Different Models focus on When Summarizing?} We compare the summary styles of TNLG v2~\citep{TNLG} (530B) to Cohere's Command X large Beta~\citep{cohere_command_2023} on the CNN/DailyMail dataset. While these models achieve a similar mean win rate on the HELM leaderboard, we see when using LLM as a preference judge, Command X has a win-rate of 71.12\%.  
Looking at the top 5 vibes located in \autoref{fig:summarization_table}, we find that (1) Command X clearly states an introduction and conclusion while TNLG utilizes choppy sentences without an either (2) Command provides specific examples or anecdotes to illustrate points and (3) Command is able to capture multiple viewpoints and emotional aspects of a story while TNLG is more objective. We see these qualities are positively correlated with human preference, which may explain the disparity between correctness metrics and preference metrics. 
With these vibes, we achieve a model matching accuracy of \textbf{71.29\%} and a preference prediction accuracy of \textbf{61.42\%}.

\begin{figure}[h]
    \centering
    \includegraphics[width=\linewidth, trim={0cm 2.4cm 0cm 0cm},clip]{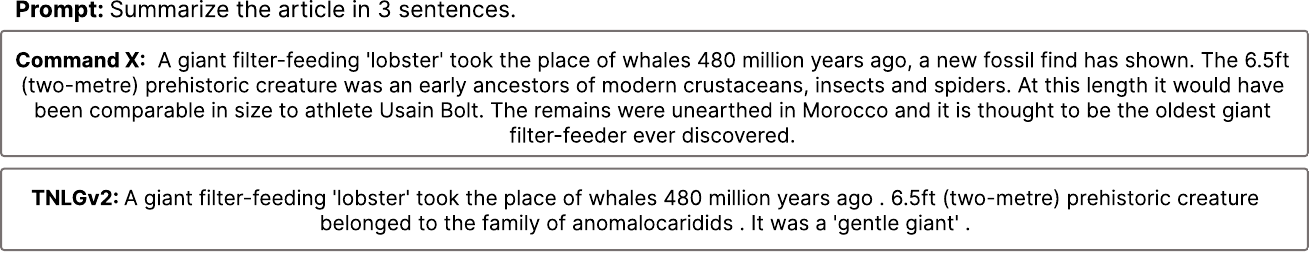}
    \label{fig:summarization}
    \vspace{-2.5em}
\end{figure}


\subsection{How do different LLMs solve math problems?} 
\begin{figure}[h]
    \centering
    \includegraphics[width=\linewidth, trim={0.4cm 23cm 0.4cm 0.5cm},clip]{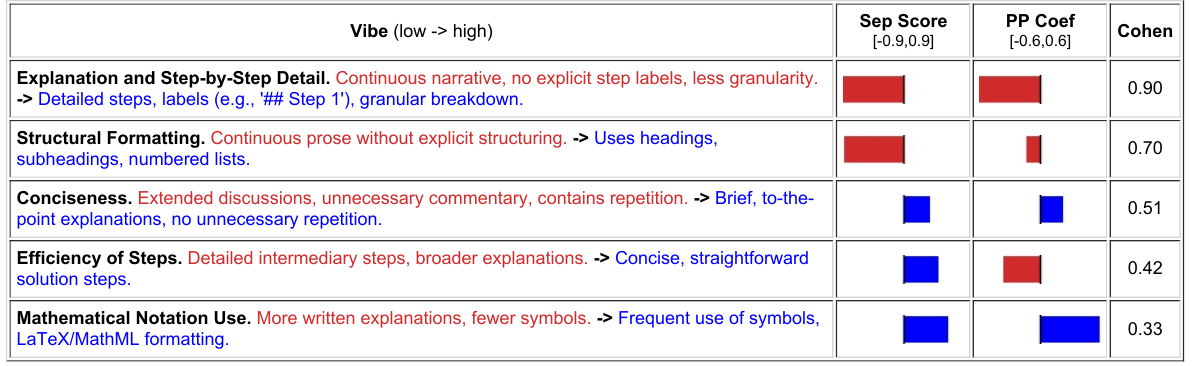}
    \vspace{-0.5cm}
    \caption{\textbf{Top 5 vibes comparing GPT-4o to Llama-3-405B on MATH CoT.} Negative separability scores indicate GPT-4o aligns with the low (red) description, while negative preference coefficients show alignment with low descriptions is preferred. GPT-4o outputs contain more LaTex/MathML formatting which positively correlated with human preference while Llama-3-405B has very structured and overly-detailed responses, which is negatively correlated with preference.}
    \vspace{-1em}
    \label{fig:math_top5_table}
\end{figure}

Objective tasks like math have a single final answer, but the way a model explains its thought process varies across models. We run \method{} on the MATH dataset~\citep{hendrycksmath2021} using chain-of-thought prompting to discover how GPT-4o and Llama-405b differ in their thought process and presentation. To reduce the variance seen from incorrect examples, we run \method{} only on the questions where both models answered correctly and aim to discover why GPT-4o is favored in 76\% of conversations. Inspecting the top 5 vibes in \autoref{fig:math_top5_table},  we observe that Llama-405b organizes its responses under markdown headings, adopts a more conversational tone, and includes overly detailed step-by-step explanations, as illustrated below. Examining the coefficients of the preference prediction model, we find that a formal tone and frequent use of notation positively correlate with preference, while over-explaining the reasoning process negatively correlates with preference.
These vibes achieve a model-matching accuracy of \textbf{97.09\%} and a preference prediction accuracy of \textbf{72.79\%}.


\begin{figure}[h]
    \centering
    \includegraphics[width=\linewidth, trim={0cm 2.4cm 0cm 0cm},clip]{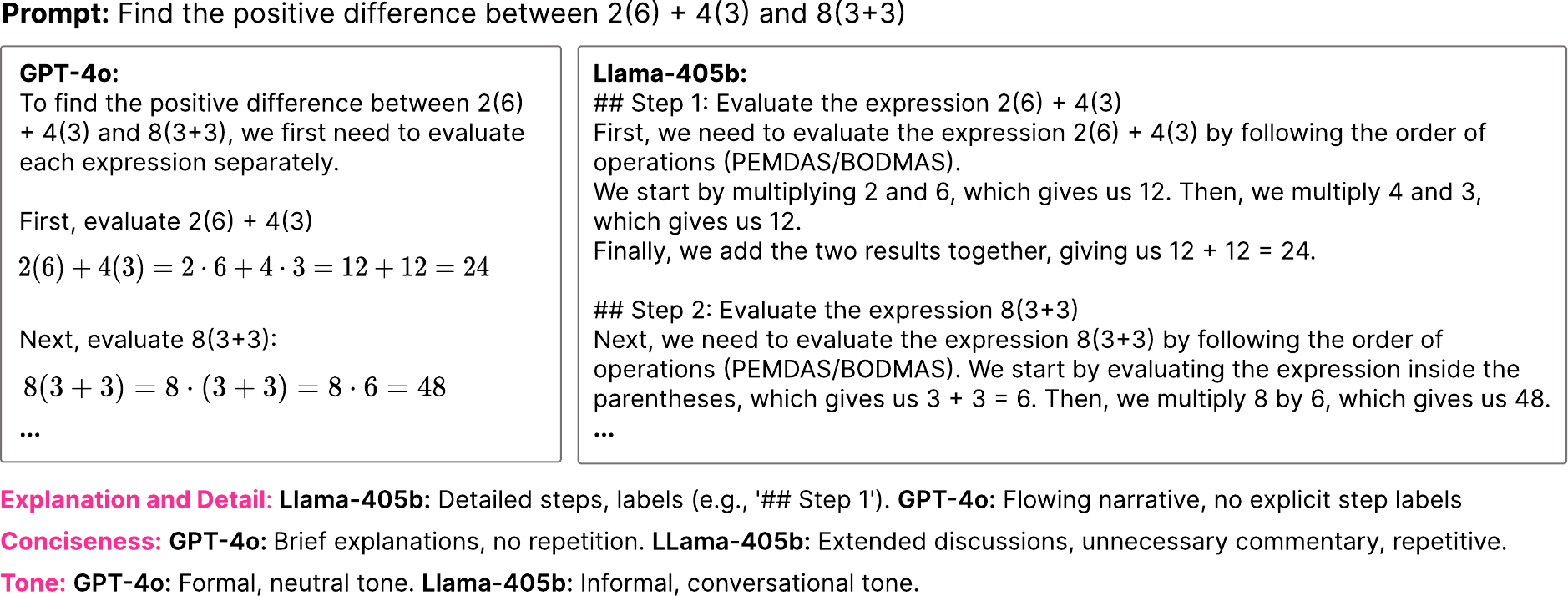}
    \vspace{-2em}
    \label{fig:math}
\end{figure}

\subsection{What are VLM's Captioning Style?}

Image captioning is one of the most popular use cases for Vision and Language models, but different captioning models focus on different image properties. We run \method{} on captions generated by GPT-4V~\citep{chen2023sharegpt4v, openai2023gpt4} and Gemini-1.5-Flash~\citep{reid2024gemini} on 1000 COCO images and we find that GPT-4V uses more poetic language and structures its captions as a dynamic story, inferring the personality and emotions of the subjects in the image while Gemini sticks to more literal descriptions (\autoref{fig:caption_vibes}). The top 10 vibes generated by \method{} are able to achieve near perfect \textbf{99.13\%} model matching accuracy and \textbf{89.02\%} preference prediction accuracy. Although we compared the captions without the image in this experiment due to cost, the \method{} framework can be easily adapted to the multimodal setting.


\begin{figure}[h]
    \centering
    \vspace{-0.3em}
\includegraphics[width=\linewidth]{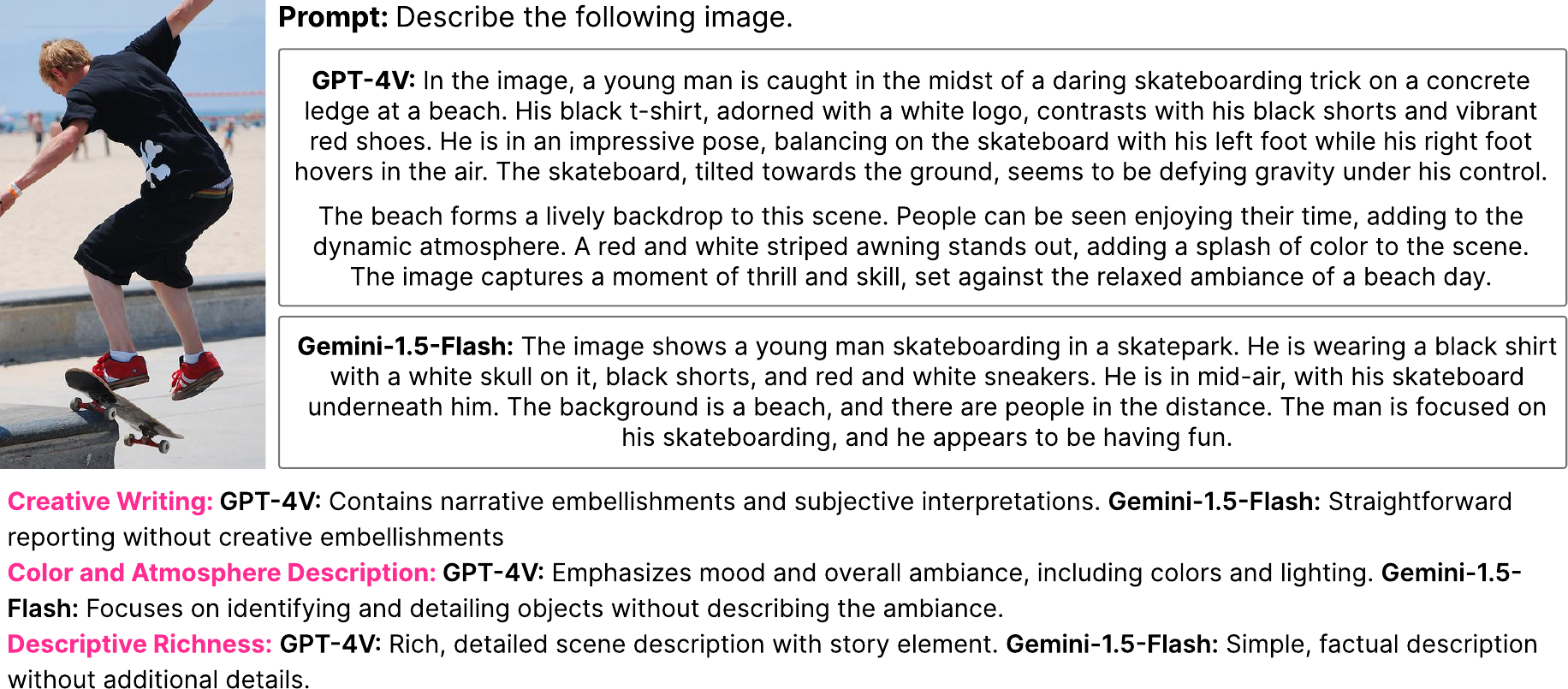}
    \vspace{-2.3em}
    \label{fig:gemini_gpt_captions}
\end{figure}

\section{Limitations}
Although \method{} quantifies the impact of each vibe on model identity and user preference, it is challenging to disentangle whether a specific vibe directly influences human preference or if other confounding factors are at play. For example, a model might exhibit a vibe of being more engaging, but its preference by users could stem from its factual accuracy, where accurate outputs incidentally appear more engaging due to their clarity or relevance. Furthermore, the LLM-based vibe discovery process may not capture all relevant differences between models. This is particularly problematic when there's a significant discrepancy in model accuracy, as the discovered vibes may focus primarily on accuracy-related aspects. \method{} is also costly to validate, as
each judge will have to evaluate each sample in $D_{validation}$ on each vibe. In order for this to be feasible, our method
uses relatively inexpensive models such as GPT-4o-mini, but these judge models are often incorrect in their predictions, as shown in \autoref{fig:math_weaknesses}. LLM judges also have biases~\citep{llmjudge}, like favoring their own outputs, which may affect the scoring. Lastly, running \method{} multiple times can lead to different vibes and different results, making it harder to reproduce findings exactly. 

\section{Conclusion}

It may seem unconventional to focus on vibes instead of concrete metrics of correctness, but these qualitative properties have a measurable impact on how people judge models. VibeCheck provides a valuable addition to existing metrics for correctness by capturing these qualitative aspects that influence human preference. As LLM usage expands, we anticipate an increased focus on evaluating vibes to better align with user preferences. Moreover, this approach can be extended to other modalities, such as audio or visual content, and can be applied to compare any pairwise set of texts, making it a versatile tool for model evaluation. In future work, we hope to explore extending this framework to compare a larger number of models along with developing interventions which can use these vibes to improve human preference for given models.  

\textbf{Acknowledgments.} We thank Ruiqi Zhong for introducing us to the joys of automated data analysis and Ion Stoica for insightful rants on evaluations beyond accuracy, as well as their feedback on the manuscript. 
We also thank Wei-Lin Chiang, Evan Frick, Tianle Li, and Issac Ong for co-authoring a blog post on the behaviors of Llama-3, which inspired one of the coolest experiments in this paper. 
Lastly, Lisa personally extends her appreciation to Joey, Jacob, and Trevor for embracing the writing of a paper that unironically uses the word "vibe" over 290 times. This paper has spawned many amusing quotes, such as: "Can we put confidence intervals on vibes?", "What if we call it `No Numbers Just Vibes', and we replace all numbers with emojis?", and of course "I'm all vibed-out".


\clearpage

\bibliography{iclr2025_conference}
\bibliographystyle{iclr2025_conference}

\appendix
\definecolor{tan}{RGB}{254, 254, 253}

\newtcolorbox{promptsupp}[1][]{
    enhanced,
    drop shadow,
    left=2mm,
    right=2mm,
    top=2mm,
    boxsep=1mm,
    sharp corners,
    boxrule=0pt,
    fontupper=\linespread{1.1}\small\ttfamily,
    colback=tan!2,
    title=#1,
}

\newtcolorbox{prompt}[1]{
    enhanced,
    drop shadow=black!5!white,
    left=4mm,
    right=4mm,
    top=3mm,
    bottom=3mm,
    boxsep=0mm,
    rounded corners,
    title=#1,
    fontupper=\linespread{1.1}\scriptsize\fontfamily{lmr}\selectfont,
    breakable
}

\section{Experimental Details \& Dataset Statistics}
\label{sec:exp_supp}

\begin{table}[h]
\centering
\begin{tabular}{lcc}
\toprule
Dataset & \# Train & \# Test \\
\midrule
Human VS ChatGPT & 250 & 250 \\
Chatbot Arena - All & 839 & 839 \\
Chatbot Arena - STEM & 346 & 347 \\
Chatbot Arena - Writing & 278 & 277 \\
CNN/DailyMail & 444 & 346 \\
MATH & 218 & 218 \\
COCO w/ ShareGPT-4V Captions & 323 & 346 \\
\bottomrule
\end{tabular}
\caption{\textbf{Dataset Statistics}}
\label{tab:dataset_stats}
\end{table}

\begin{table}[h]
\centering
\begin{tabular}{lccc}
\toprule
Dataset & Model A & Model B & Model A Win Rate\\
\midrule
Human VS ChatGPT & Humans & GPT-3.5 & - \\
Chatbot Arena - All & Llama3-70b-Instruct & GPT-4 + Claude-3-Opus & 50\%\\
Chatbot Arena - STEM & Llama3-70b-Instruct & GPT-4 + Claude-3-Opus & 44\%\\
Chatbot Arena - Writing & Llama3-70b-Instruct & GPT-4 + Claude-3-Opus & 57\%\\
CNN/DailyMail & Cohere Command X & TNLGv2 & 71.12\%\\
MATH & GPT-4o & Llama3-405b & 76\%\\
COCO w/ ShareGPT-4V Captions & GPT-4V & Gemini-1.5-Flash & 80\%\\
\bottomrule
\end{tabular}
\caption{\textbf{Model Win Rates}}
\label{tab:model_win_rates}
\end{table}

\begin{table}[h]
\centering
\begin{tabular}{lccccc}
\toprule
Dataset & $d$ & $batch$ & $num\_eval\_vibes$ & $num\_final\_vibes$ & $iterations$\\
\midrule
Human VS ChatGPT & 40 & 5 & 10 & 10 & 3\\
Chatbot Arena - All & 20 & 5 & 10 & 10 & 3\\
Chatbot Arena - STEM & 20 & 5 & 10 & 10 & 3\\
Chatbot Arena - Writing & 20 & 5 & 10 & 10 & 3\\
CNN/DailyMail & 20 & 2 & 10 & 10 & 3\\
MATH & 20 & 5 & 10 & 10 & 1\\
COCO & 20 & 5 & 10 & 10 & 1 \\
\bottomrule
\end{tabular}
\caption{\textbf{\method{} Hyperparameters}}
\label{tab:hyperparameters}
\end{table}

$num\_eval\_vibes$ = number of vibes to validate at every iteration

$d$ = number of prompt output triples to use in each iteration of the vibe discovery phase

$batch$ = number of triples to feed into the prompt of the discovery LLM at once. 

$iterations$ = number of vibe iterations to perform 

$num\_final\_vibes$ = number of vibes to evaluate at the end of all the iterations. This can be set to false, in which case all the vibes collected in the iteration

We take the 1000 captions generated by GPT-4V from the ShareGPT-4V dataset~\cite{chen2023sharegpt4v} and generate captions for the same images using the same captioning prompt using Gemini-1.5-Flash. 

\clearpage

\section{Gold standard labels}
\label{sec:gold_standard}
Below is a summary of key differences found by human evaluators in the HC3 dataset~\cite{guo-etal-2023-hc3} listed in their paper. 

\textbf{Characteristics of ChatGPT}

\begin{enumerate}[label=(\alph*)]
    \item Responses are well-organized, often starting with a definition of key concepts before providing a step-by-step explanation and concluding with a summary.
    \item Answers tend to be detailed and extensive.
    \item ChatGPT generally minimizes bias and avoids generating harmful content.
    \item It refrains from responding to queries beyond its scope of knowledge.
    \item In some cases, it may generate incorrect or fabricated information.
\end{enumerate}

\textbf{Differences Between Human and ChatGPT Responses}

\begin{enumerate}[label=(\alph*)]
    \item ChatGPT remains strictly on topic, while human responses may shift toward related or tangential subjects.
    \item It tends to provide objective, fact-based answers, whereas human responses often include personal opinions or subjective elements.
    \item ChatGPT's tone is typically formal and structured, while human speech is more conversational and informal.
    \item Unlike humans, ChatGPT does not express emotions, relying solely on linguistic structure rather than emotional cues like punctuation or tone variations.
\end{enumerate}

\section{Generating Preset Vibes}
\label{sec:preset_vibes}

\begin{table}[h]
\setlength\tabcolsep{1pt} 
\centering
\adjustbox{max width=\textwidth}{
\begin{tabularx}{\textwidth}{L{2.5cm} L{11.5cm}}
\toprule
\textbf{Vibe} & \textbf{Axis Definition (low $\to$ high}) \\
\midrule
Assertiveness & 
Uses tentative or uncertain language. $\to$ Uses definitive, confident statements. \\
\addlinespace[1pt]
Detail \& Elaboration & 
Gives brief or shallow responses. $\to$ Provides thorough, nuanced, and expansive information. \\
\addlinespace[1pt]
Formality & 
casual, conversational, or informal language. $\to$ formal, sophisticated language and sentence structure. \\
\addlinespace[1pt]
Emotional Tone & 
Remains neutral or detached. $\to$ Infuses responses with expressive emotion and enthusiastic or empathetic tone. \\
\addlinespace[1pt]
Creativity \& Originality & 
Sticks to standard, predictable answers. $\to$ Provides responses with novel ideas or imaginative scenarios. \\
\addlinespace[1pt]
Explicitness & 
Uses vague or implicit language. $\to$ States things directly and unambiguously. \\
\addlinespace[1pt]
Humor and Playfulness & 
Responds in a straightforward and serious manner. $\to$ Uses humor, playful language, or wordplay. \\
\addlinespace[1pt]
Engagement & 
Presents information passively. $\to$ Actively engages the reader using rhetorical questions or interactive phrasing. \\
\addlinespace[1pt]
Logical Rigor & 
Provides conclusions without thorough justification. $\to$ Constructs well-supported arguments with clear reasoning. \\
\addlinespace[1pt]
Conciseness & 
Uses verbose language and excessive details. $\to$ Uses minimal words to convey a point clearly. \\
\bottomrule
\end{tabularx}
}
\caption{\textbf{Predefined vibes.} We prompt GPT-4o to generate a set of 10 vibes which represent common axes on which LLM outputs differ. }
\label{tab:axis_definitions}
\end{table}

We generate our list of 10 preset vibes by prompting GPT-4o with the following:

\begin{promptsupp}[Preset Vibe Generation Prompt]
I am a machine learning researcher trying to figure out the major differences between the behavior of different large language models. Can you list common ways in which two language models can differ in their outputs? \\
        
Please output a list differences between these sets of outputs with relation to specific axes of variation. Try to give axes that a human could easily interpret and they could understand what it means to be higher or lower on that specific axis. Please ensure that the concepts used to explain what is high and low on the axis are distinct and mutually exclusive such that given any tuple of text outputs, a human could easily and reliably determine which model is higher or lower on that axis. \\

The format should be \\
- \{{axis 1}\}: \{{difference}\} \\
- \{{axis 2}\}: \{{difference}\} \\
    
Please output differences which have a possibility of showing up in future unseen data and which would be useful for a human to know about when deciding with LLM to use. For each axis, define clearly and succinctly what constitutes a high or low score, ensuring these definitions are mutually exclusive. Please give 10 differences
\end{promptsupp}

\section{Additional \method{} Details}
\label{sec:supp_method}

\subsection{Vibe Discovery}

Below is the user prompt we use for vibe discovery. 
\begin{promptsupp}[Vibe Discovery Prompt]
The following are the results of asking a set language models to generate an answer for the same questions:

[PROMPT]
[OUTPUT 1]
[OUTPUT 2]

I am a machine learning researcher trying to figure out the major differences between these two LLM outputs so I can better compare the behavior of these models. Are there any variations you notice in the outputs? 

Please output a list differences between these sets of outputs with relation to specific axes of variation. Try to give axes that a human could easily interpret and they could understand what it means to be higher or lower on that specific axis. Please ensure that the concepts used to explain what is high and low on the axis are distinct and mutually exclusive such that given any tuple of text outputs, a human could easily and reliably determine which model is higher or lower on that axis.

The format should be: \{\{axis\}\}: Low: \{\{low description\}\}; High: \{\{high description\}\}
\end{promptsupp}

\textbf{Vibe Summarization.} To summarize the set of vibes found in the vibe discovery process, We cluster the axes using agglomerative clustering on the embeddings of the axes generated by the 'hkunlp/instructor-xl' model, and prompt GPT-4o to reduce this set by removing any vibes which are similar. After this stage we are left with a set of less than 20 vibes which we use to score the outputs of each model.

\begin{promptsupp}[Vibe Reduction Prompt]
    Below is a list of axes with a description of what makes a piece of text low or high on this axis. Are there any axes that have similar meanings based off their low and high descriptions? Are there any sets of axes that would convey the same information to a user (e.g. level of detail)? Could any of the low and high descriptions be simplified to make them easier to understand?
    
Please remove any axes with roughly the same meaning and simplify the descriptions of what makes a piece of text low or high on this axis. Please ensure that the descriptions of what makes a piece of text low or high on this axis are distinct, useful, and mutually exclusive. Given any piece of text, a human should be able to easily and reliably determine if this text falls high or low on each axis. 

Here is the list of axes:
\{axes\} \\

Please return the simplified list of axes and the descriptions of what makes a piece of text low or high on this axis. These axes should contain only one concept and should be human interpretable. 
Some examples of bad axes include:

- "Configuration Clarity: High: Clearly defined structure and purpose. Low: Vaguely defined, minimal purpose." -> This axes is bad because it is not clear what a clearly defined purpose means nor what a vaugely defined purpose means. 

- "Language and Communication: High: Varied/precise, complex structure. Low: Straightforward, simple or general language." -> This axes is bad because it combines multiple concepts into one axis.

- "Content Quality: High: High quality, engaging, informative. Low: Low quality, unengaging, uninformative." -> This axes is bad because it is not clear what high quality means nor what low quality means. \\

    Some examples of good axes include: 
    
- "Complexity: High: Complex, multi-layered, intricate. Low: Simple, straightforward, easy to understand."

- "Efficiency (coding): High: Code optimized for runtime, minimal memory usage. Low: Code inefficient, high memory usage." \\

Some examples of axes which should be combined include:

- "Emotional Tone: High: Contains emotionally charged language. Low: Maintains a neutral tone." and "Empathy: High: Shows empathy. Low: Only factual answers without empathy." are redundant because they both measure the emotional content of the text. If two similar axes are found, keep the one that is more informative or more specific.\\

Please maintain the format of the original axes and return a list like ["\{{axis name}\}: High: \{{high description}\} Low: \{{low description}\}", ...]. I should be able to parse this output into a string using $\texttt{ast.literal\_eval}$. If the original list does not contain any redundant axes, please return the original list.
\end{promptsupp}

If the number of vibes after the first reduction step is $>K$, we prompt GPT-4o to reduce the set further with the final reducer prompt.

\begin{promptsupp}[Final Vibe Reducer Prompt]
    Below is a list of axes with a description of what makes a piece of text low or high on this axis. I would like to summarize this list to at most {number} representative axes. \\

    Here is the list of axes:
    [VIBES] \\

    These axes should contain only one concept and should be human interpretable. Some examples of bad axes include: \\
    - "Configuration Clarity: High: Clearly defined structure and purpose. Low: Vaguely defined, minimal purpose." -> This axis is bad because it is not clear what a clearly defined purpose means nor what a vaguely defined purpose means.  \\
    - "Language and Communication: High: Varied/precise, complex structure. Low: Straightforward, simple or general language." -> This axis is bad because it combines multiple concepts into one axis. \\
    - "Content Quality: High: High quality, engaging, informative. Low: Low quality, unengaging, uninformative." -> This axis is bad because it is not clear what high quality means nor what low quality means.

    Some examples of good axes include: \\
    - "Complexity: High: Complex, multi-layered, intricate. Low: Simple, straightforward, easy to understand." \\
    - "Efficiency (coding): High: Code optimized for runtime, minimal memory usage. Low: Code inefficient, high memory usage." \\

    Some examples of axes which should be combined include: \\
    - "Emotional Tone: High: Contains emotionally charged language. Low: Maintains a neutral tone." and "Empathy: High: Shows empathy. Low: Only factual answers without empathy." are redundant because they both measure the emotional content of the text. If two similar axes are found, keep the one that is more informative or more specific.

    Please return the simplified list of <=[K] axes with any redundant axes removed and the descriptions of what makes a piece of text low or high on this axis simplified. Are there any axes which convey roughly the same information? Are there any axes where almost all samples which score highly on one axis would also score highly on the other? \\

    Please maintain the format of the original axes and return a numbered list. Each element should be structured as follows:
    "\{{axis name}\}: High: \{{high description}\} Low: \{{low description}\}"
\end{promptsupp}

\subsection{Vibe Validation}
\begin{promptsupp}[Prompt for ranker judge]
I want to compare the outputs of two language models (A and B) for the same prompt. I would like you to evaluate where each output falls on the following axis: [VIBE]. 

If you had to choose which output is higher on the axis, which would you choose? Here is the prompt and the outputs of A and B respectively:

[PROMPT][OUTPUT A][OUTPUT B]

Please respond with which model you think is higher on the axis and explain your reasoning. If this axis does not apply to these examples or these outputs are roughly equal on this axis, return "N/A".
\end{promptsupp}


\subsection{Vibe Iteration}

At iteration step $t$, we are left with $k$ distinct vibes which are well-defined and differentiating along with their scores $\vibe_{1:k}(\pr, \out_A, \out_B)$. Using these scores, we train a LR model to predict LLM identity (i.e. "Is the response shown first LLM A or LLM B?") and get the predictions on our entire set $D$. Assuming we have not hit the max iteration steps set by the user, we iterate if the number of samples misclassified by the model matching predictor is greater than the number of prompts to perform discovery on ($d$). In iteration step $t+1$, we take these misclassified prompt output triples in batches of size $batch$ along with the current set of vibes $\vibe_1, ..., \vibe_k$ and prompt the LLM to generate new differences between outputs what are not represented in the current vibes. These vibes are then reduced using the same procedure as the vibe discovery process. In practice we found that often some of the reduced vibes from the discovery phase at $t+1$ were redundant with an existing axis, so we preform one more deduplication step using the prompt below. 

\begin{promptsupp}[Vibe Discovery Iteration step]

Given a new set of respenses, your task is to expand on the set of axes which have been previously identified by finding other clear differences between the responses that are not captured by the existing axes. The expanded axes should be any differences between responses that are not clearly captured by the existing axes. Be as exhaustive as possible in listing differences on as many different axes as you can think of, and be specific about what constitutes high and low on each axis. \\

Your axis should be interpretable: a human should easily and reliably determine which response is higher, lower, or even on this axis when given a new set of responses. Please do not make your axes too broad and list as many axes as you can think of that are not covered by the existing axes. Most of these new axes should be either completely different from the existing axes or should highlight a more finegrained difference which an existing axis might broadly cover. For instance, if an existing axis is "Enthusiasm: High: enthusiastic, Low: unenthusiastic", a new axis might be "Use of Exclamation Points", or if an existing axis is "Cultural Context: High: culturally relevant, Low: culturally irrelevant", a new axis might be "Use of Slang".
", a new axis might be "Use of Exclamation Points", or if an existing axis is "Context", a new axis might be "". \\

Please think through the axes carefully and make sure they are clear, concise, and do not overlap with eachother or the existing axes. Do not include any of the existing axes in your response. Your output should be in this format: \\

New Axes: \\
- {{axis 1}}: \\
    High: {{description of high}} \\
    Low: {{description of low}} \\

- {{axis 2}}: \\
    High: {{description of high}} \\
    Low: {{description of low}} \\

Do not include any other information in your response.
\end{promptsupp}

\begin{promptsupp}[Vibe deduplication in iteration step $t+1$]
Here is a list of axes on which two strings may vary. Each axis has a description of what makes a string high or low on that axis. \\

[EXISTING AXES] 

[NEW AXES] \\

It is likely that several of these axes measure similar things. Your task is to remove any redundant axes. Think about if a user would gain any new information from seeing both axes. For example, "Emotional Tone: High: Contains emotionally charged language. Low: Maintains a neutral tone." and "Empathy: High: Shows empathy. Low: Only factual answers without empathy." are redundant because they both measure the emotional content of the text. If two similar axes are found, keep the one that is more informative. \\

Output the reduced list of axes, separated by a newline. All of the axes should maintain the same format they have in the list of \{{axis}\}: High: \{{high}\} Low: \{{low}\}
\end{promptsupp}

\subsection{Generating Preference Labels}
\begin{promptsupp}[prompt for generating preference labels]
    Please act as an impartial judge and evaluate the quality of the responses provided by two AI assistants (A and B) to the user question displayed below. You should choose the assistant that follows the user’s instructions and answers the user’s question better. Your evaluation should consider factors such as the helpfulness, relevance, accuracy, depth, creativity, and level of detail of their responses. Begin your evaluation by comparing the two responses and provide a short explanation. Avoid any position biases and ensure that the order in which the responses were presented does not influence your decision. Do not allow the length of the responses to influence your evaluation. Do not favor certain names of the assistants. Be as objective as possible. \\

    Here is the prompt and the outputs of A and B respectively:

    [PROMPT][OUTPUT A][OUTPUT B] \\

    Please respond with the model which contains a higher quality response. Based on your analysis, please explain your reasoning before assigning a score. Use the following format for your response:
    
        Analysis: \{{reasoning}\}
        
        Model: \{{A, B, tie}\}
\end{promptsupp}

\section{Further Related Works}
\label{sec:more_related}

\textbf{Automatic metrics for benchmark evaluations.} The number of benchmarks in the NLP community has exploded in recent years, with a wealth of work on providing a more holistic evaluation of language models beyond just accuracy. Several works \cite{pang2020towards, banerjee-lavie-2005-meteor, bluert},  aim to improve on automatic metrics like BLEU~\cite{papineni2002bleu} and ROUGE~\cite{lin-2004-rouge} scores to better measure how well a models output aligns with the ground truth by incorporating more nuanced evaluation criteria like factual accuracy, fluency, and conciseness. Similarly, efforts have been made~\cite{helm} to standardize model evaluation by evaluating models on many of these metrics.   

\newpage 

\section{Limitations}

\begin{figure}[H]
    \centering
    \includegraphics[width=\linewidth]{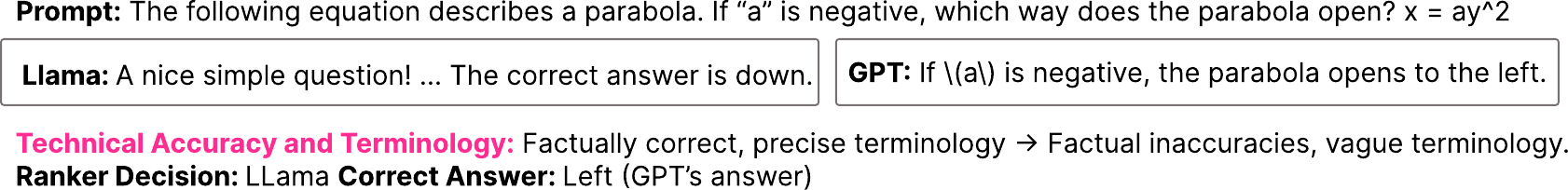}
    \caption{\textbf{Weaknesses in the mathematical abilities of the LLM judge (GPT-4o-mini).}}
    \label{fig:math_weaknesses}
\end{figure}

\begin{figure}[h]
    \centering
    \includegraphics[width=\linewidth]{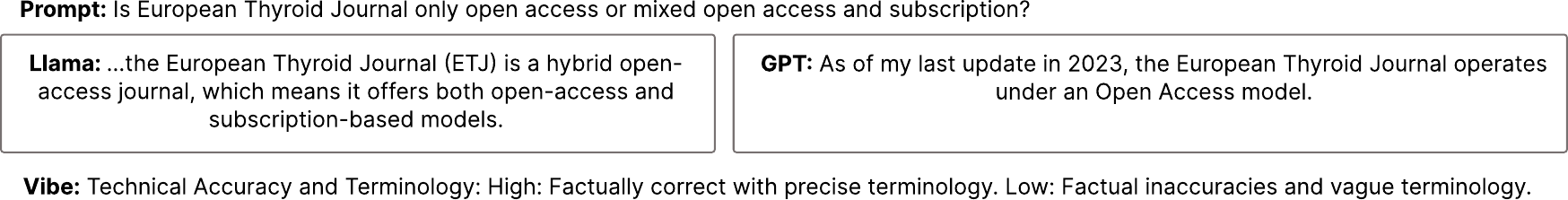}
    \caption{The answer to certain questions changes depending on the following parameters: \\ 
    (1) When was the question asked?\\
    (2) What is the knowledge cutoff of Model A and Model B?\\
    (3) What is the knowledge cutoff of the LLM ranker ensemble?\\
    These types of questions lead to unreliable ranker evaluations and reduced inter-annotator agreement.}
    \label{fig:time_based_questions}
\end{figure}

\section{Vibes from each Application}
\label{sec:supp_vibes}

\begin{figure}[h]
    \centering
    \includegraphics[width=\linewidth, trim={0.4cm 18cm 0.4cm 0.5cm},clip]{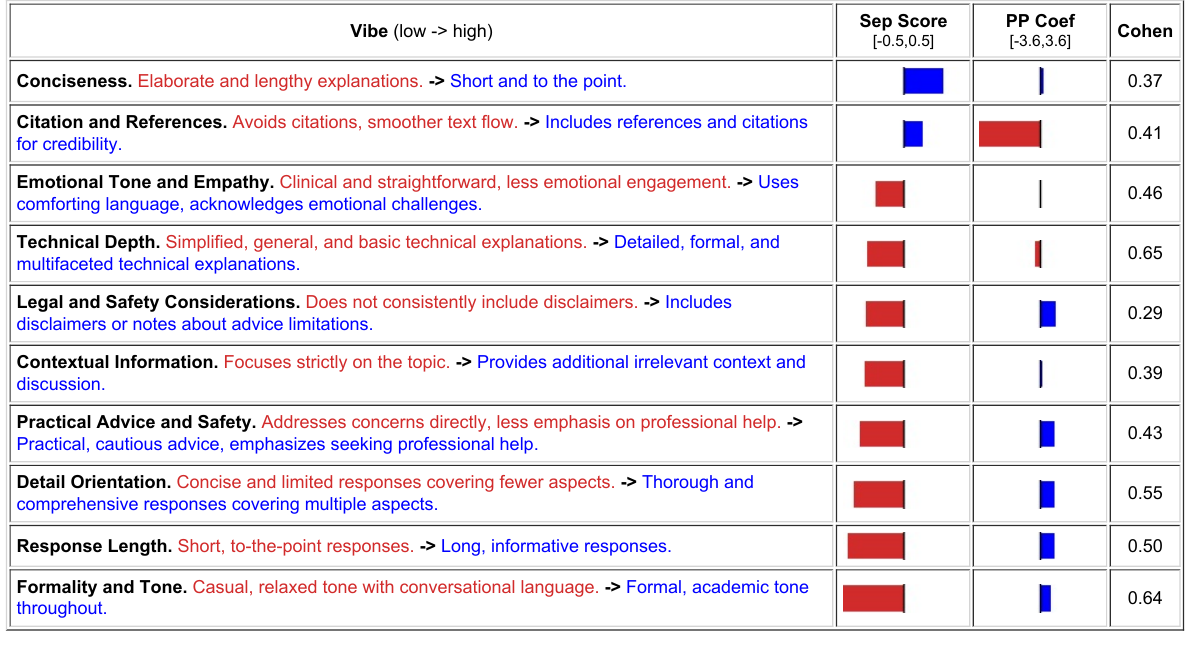}
    \caption{Human VS ChatGPT outputs on HC3~\citep{guo-etal-2023-hc3}}
    \label{fig:hec3_table_overall}
\end{figure}

\begin{figure}[h]
    \centering
    \includegraphics[width=\linewidth, trim={0.4cm 18.2cm 0.4cm 0.5cm},clip]{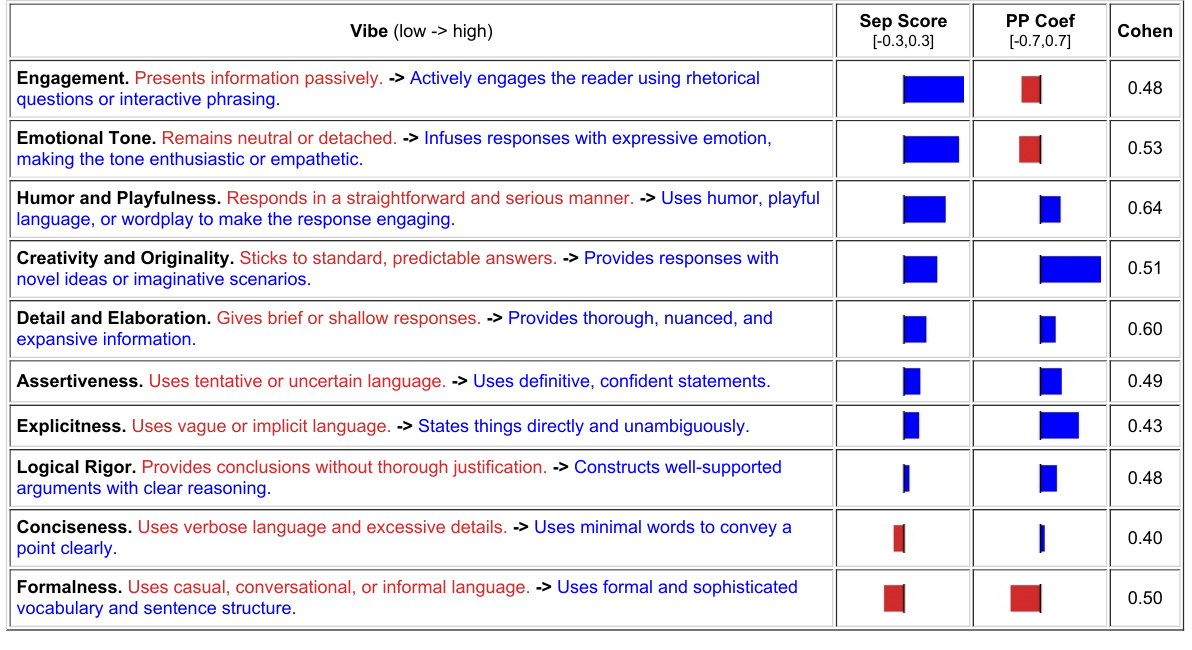}
    \caption{Preset vibes on Chatbot Arena[Overall]}
    \label{fig:preset_table_overall}
\end{figure}

\begin{figure}[h]
    \centering
    \includegraphics[width=\linewidth, trim={0.4cm 18.2cm 0.4cm 0.5cm},clip]{figures/vibe_tables/arena_all.pdf}
    \caption{VibeCheck vibes on Chatbot Arena[Overall]}
    \label{fig:preset_table_overall}
\end{figure}

\begin{figure}[h]
    \centering
    \includegraphics[width=\linewidth, trim={0.4cm 18.2cm 0.4cm 0.5cm},clip]{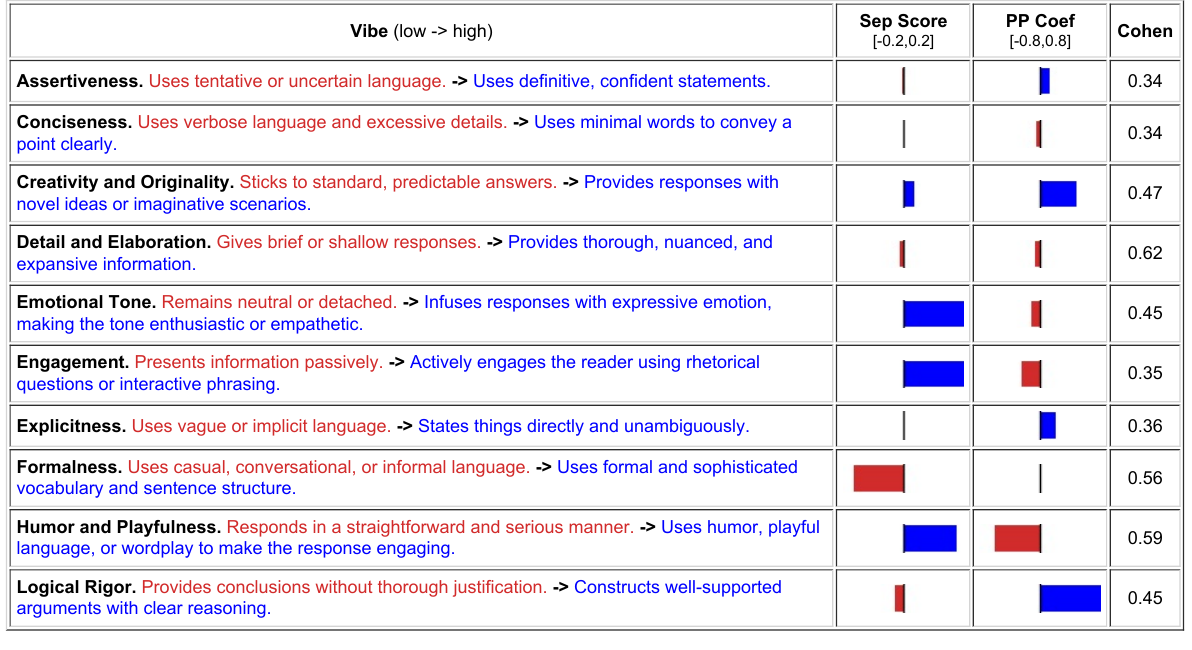}
    \caption{Preset vibes on Chatbot Arena[STEM]}
    \label{fig:preset_table_stem}
\end{figure}

\begin{figure}[h]
    \centering
    \includegraphics[width=\linewidth, trim={0.4cm 20.8cm 0.4cm 0.5cm},clip]{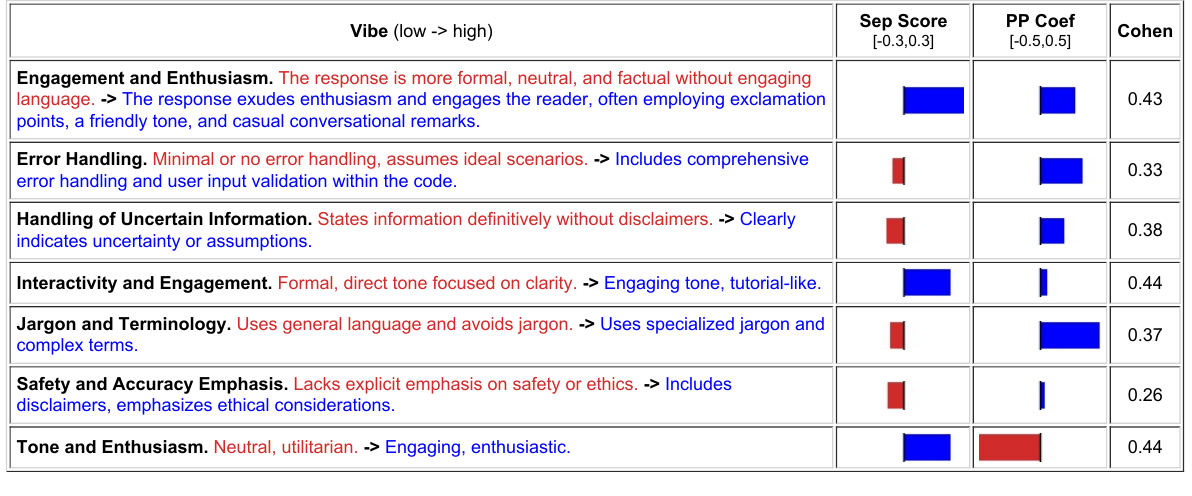}
    \caption{VibeCheck vibes on Chatbot Arena [STEM]. Note that we only find 7 vibes which achieve a separability score on the training set about the 0.05 threshold.}
    \label{fig:arena_stem_3_iter_table}
\end{figure}

\begin{figure}[h]
    \centering
    \includegraphics[width=\linewidth, trim={0.4cm 18.4cm 0.4cm 0.5cm},clip]{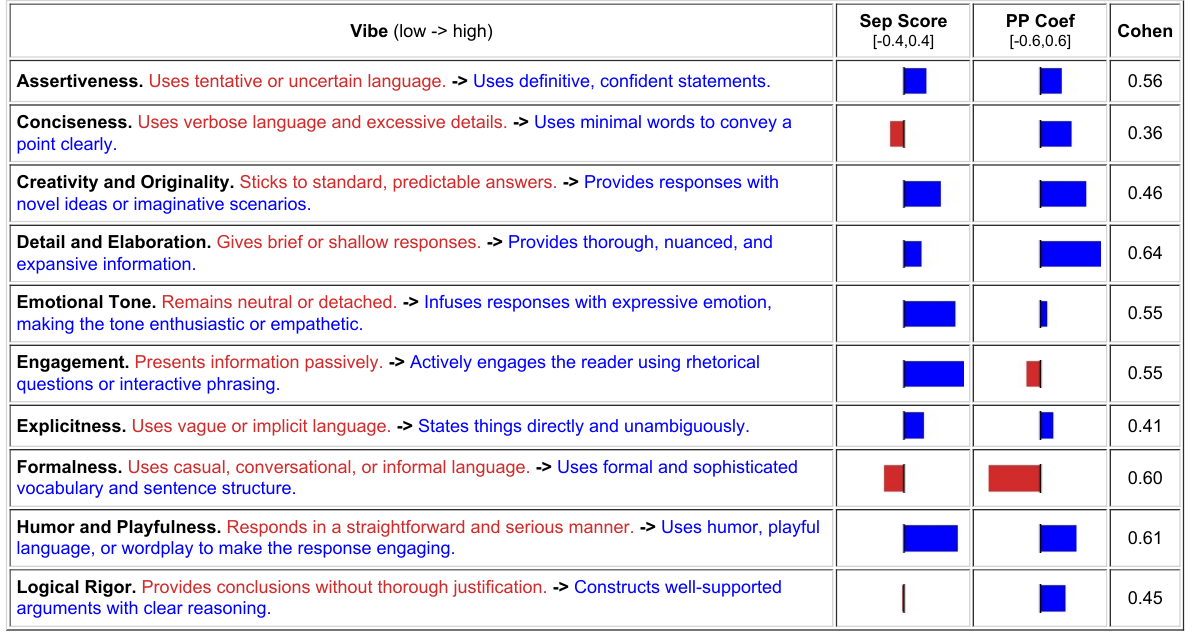}
    \caption{Preset vibes on Chatbot Arena [Writing]}
    \label{fig:preset_table_writing}
\end{figure}

\begin{figure}[h]
    \centering
    \includegraphics[width=\linewidth, trim={0.4cm 18.8cm 0.4cm 0.5cm},clip]{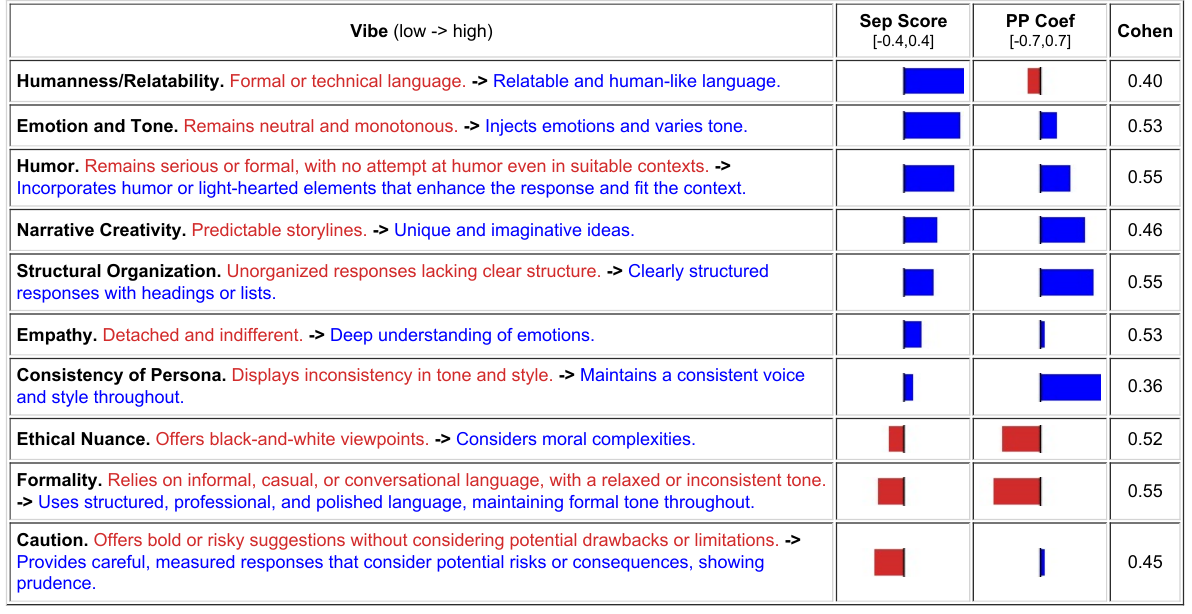}
    \caption{VibeCheck vibes on Chatbot Arena [Writing]}
    \label{fig:arena_writing_3_iter_table}
\end{figure}

\begin{figure}[h]
    \centering
    \includegraphics[width=\linewidth, trim={0.4cm 18cm 0.4cm 0.5cm},clip]{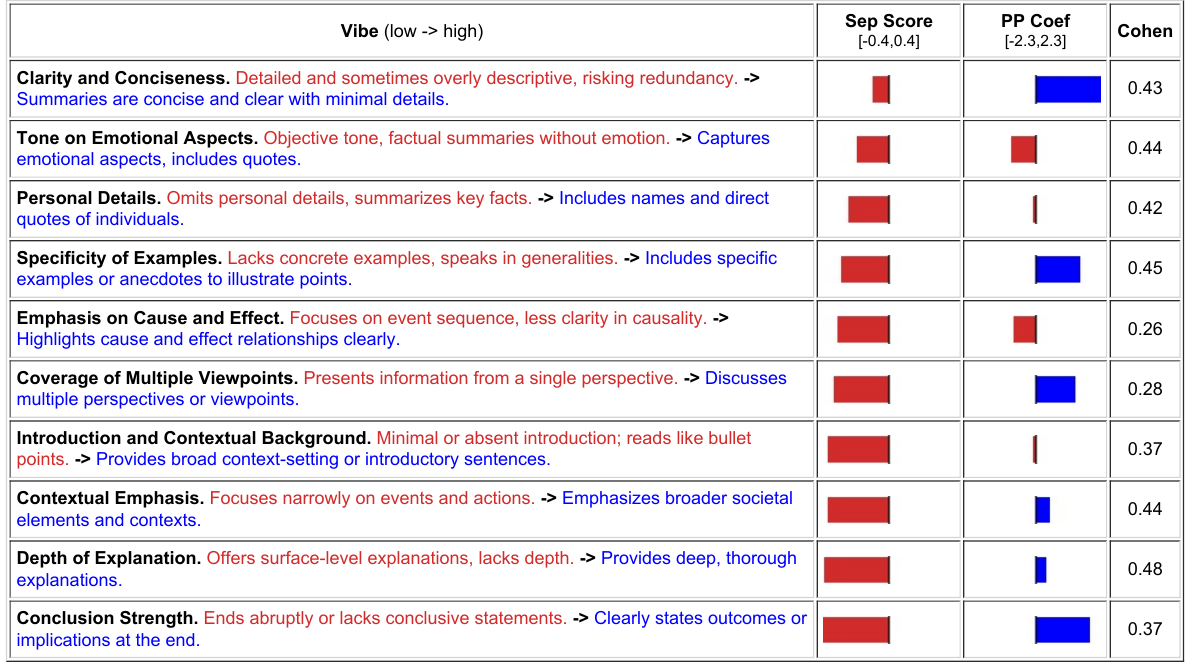}
    \caption{VibeCheck vibes comparing TNLGv2 to Command X Large Beta on CNN/DailyMail Summarization~\citep{cnn_dm}.}
    \label{fig:summarization_table}
\end{figure}

\begin{figure}[h]
    \centering
    \includegraphics[width=\linewidth, trim={0.4cm 23cm 0.4cm 0.5cm},clip]{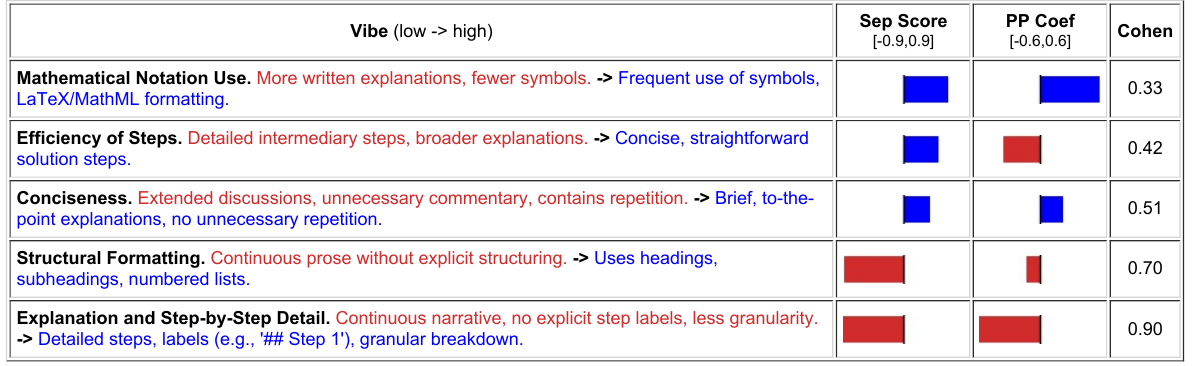}
    \caption{VibeCheck vibes comparing GPT-4o to Llama-3-405B on MATH CoT~\citep{hendrycksmath2021}. We only find 5 vibes because the vibe reduction step is not required to return $\leq$ 10 vibes and in this case found only 5 distinct vibes which are able to almost perfectly separate model outputs.}
    \label{fig:math_cot_vibes}
\end{figure}

\begin{figure}[h]
    \centering
    \includegraphics[width=\linewidth, trim={0.4cm 17.5cm 0.4cm 0.5cm},clip]{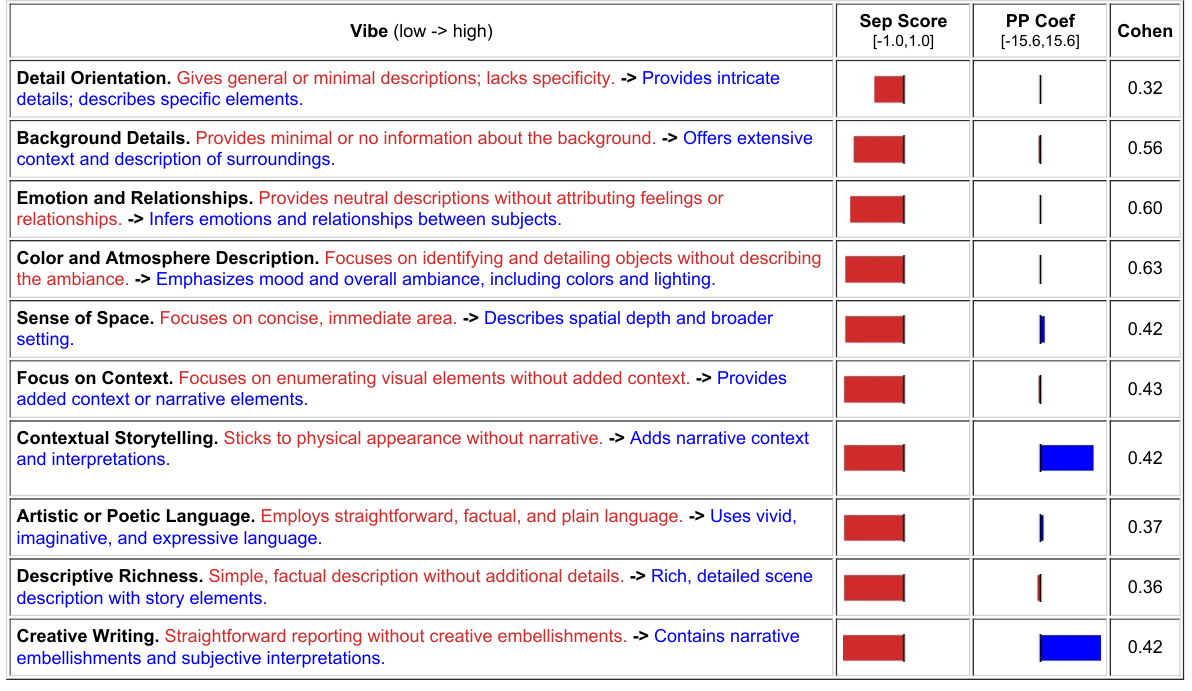}
    \caption{VibeCheck vibes comparing Gemini-1.5-Flash to GPT-4V on COCO Captions~\citep{coco}.}
    \label{fig:caption_vibes}
\end{figure}

\clearpage

\section{More LLama VS GPT Examples}
\label{sec:more_examples}

\begin{figure}[h]
    \centering
    \includegraphics[width=0.8\linewidth]{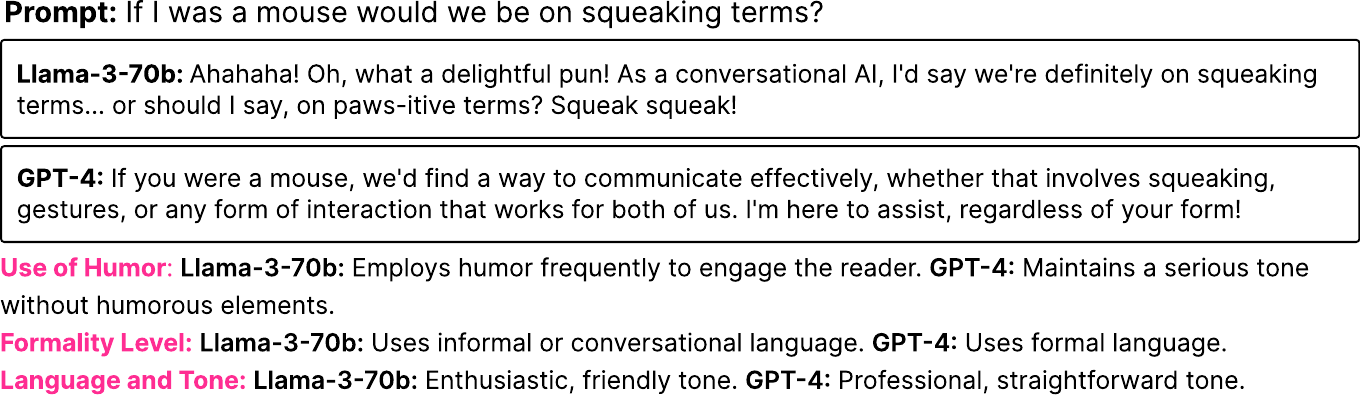}
    \caption{Squirrel example from Chatbot Arena}
    \label{fig:llama_example_main}
\end{figure}

\begin{figure}[h]
    \centering
    \includegraphics[width=\linewidth]{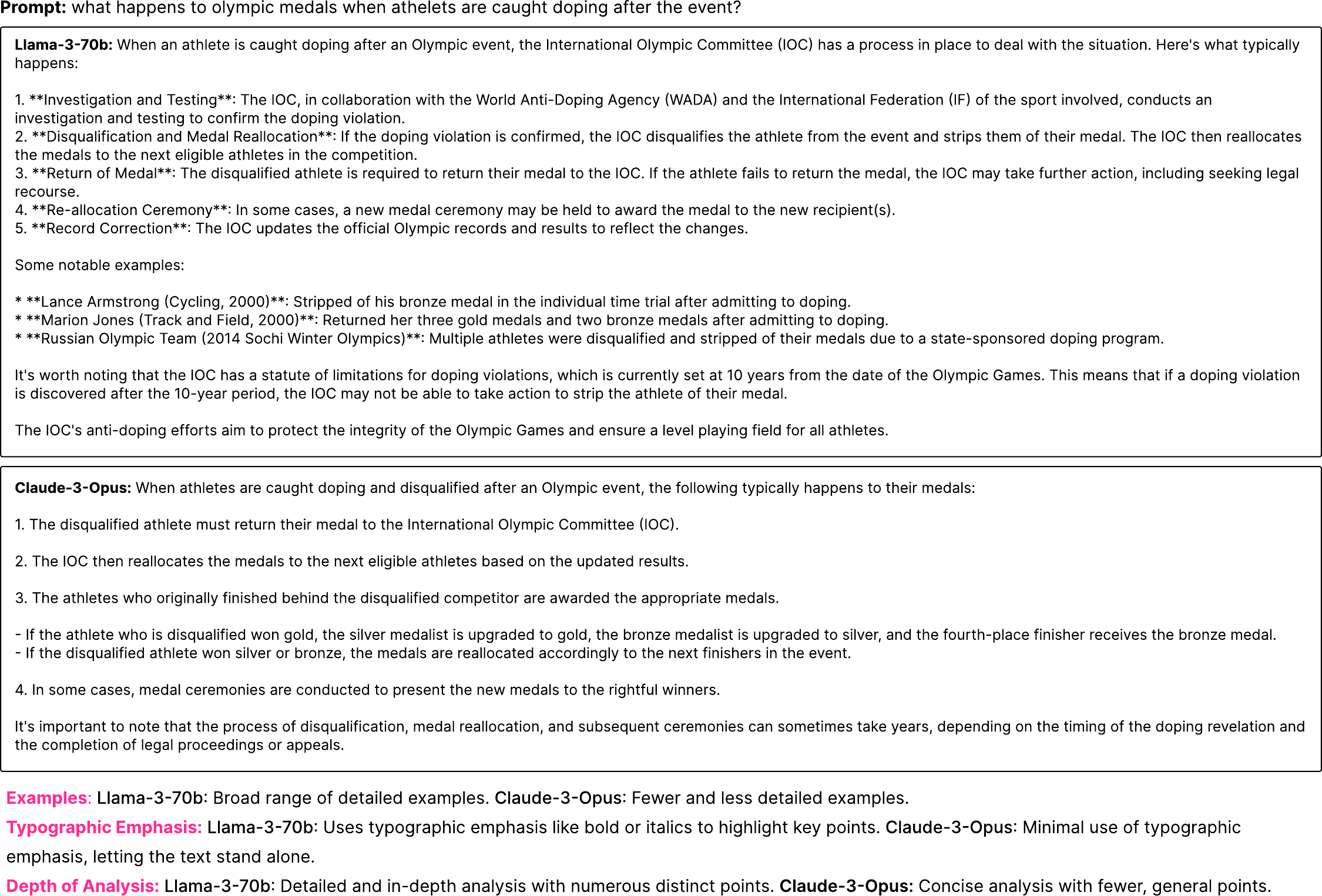}
    \caption{Olympics example from Chatbot Arena}
    \label{fig:olympics}
\end{figure}

\begin{figure}[h]
    \centering
    \includegraphics[width=\linewidth]{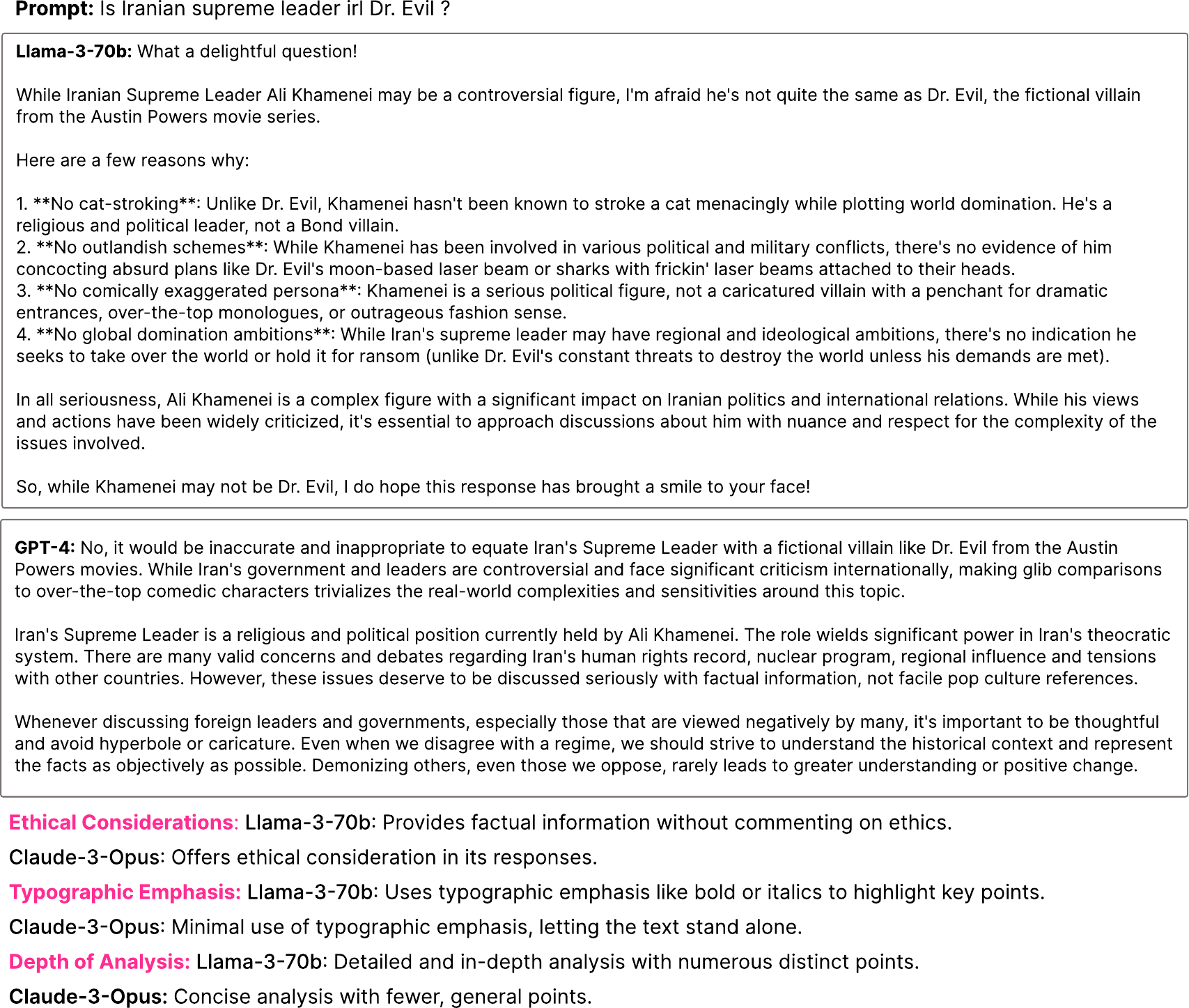}
    \caption{Supreme Leader example from Chatbot Arena}
    \label{fig:supreme_leader}
\end{figure}

\end{document}